\documentclass[graybox]{svmult}

\usepackage{type1cm}        
\usepackage{makeidx}         
\usepackage{graphicx}        
\usepackage{multicol}        
\usepackage[bottom]{footmisc}

\usepackage{amsmath}
\usepackage{amssymb}
\usepackage{booktabs}
\usepackage{multirow}
\usepackage{times}
\usepackage{setspace}
\usepackage{subfigure}
\usepackage{amsfonts}
\usepackage{comment}
\usepackage{color}
\usepackage{tabularx}
\usepackage{makecell}
\usepackage{enumerate}
\usepackage{hyperref}
\usepackage{dsfont}

\DeclareMathOperator*{\argmin}{min}

\makeindex             

\begin{document}

\title*{Computational Emotion Analysis From Images: Recent Advances and Future Directions}
\titlerunning{Computational Emotion Analysis From Images}
\author{Sicheng Zhao, Quanwei Huang, Youbao Tang, Xingxu Yao, Jufeng Yang, Guiguang Ding, Bj{\"o}rn W.\ Schuller}
\authorrunning{Sicheng Zhao et al.}
\institute{Sicheng Zhao, Quanwei Huang, Guiduang Ding \at Tsinghua University \email{schzhao@gmail.com}
\and Youbao Tang \at PAII Inc.
\and Xingxu Yao, Jufeng Yang \at Nankai University
\and Bj{\"o}rn W. Schuller \at GLAM, Imperial College London, UK
}
%
%
\maketitle


\abstract{Emotions are usually evoked in humans by images. Recently, extensive research efforts have been dedicated to understanding the emotions of images. In this chapter, we aim to introduce image emotion analysis (IEA) from a computational perspective with the focus on summarizing recent advances and suggesting future directions. We begin with commonly used emotion representation models from psychology. We then define the key computational problems that the researchers have been trying to solve and provide supervised frameworks that are generally used for different IEA tasks. After the introduction of major challenges in IEA, we present some representative methods on emotion feature extraction, supervised classifier learning, and domain adaptation. Furthermore, we introduce available datasets for evaluation and summarize some main results. Finally, we discuss some open questions and future directions that researchers can pursue.
}

\section{Introduction}
\label{Sec:Introduction}

\begin{figure}[!t]
\begin{center}
\subfigure[Fear]{
\includegraphics[width=0.4\textwidth]{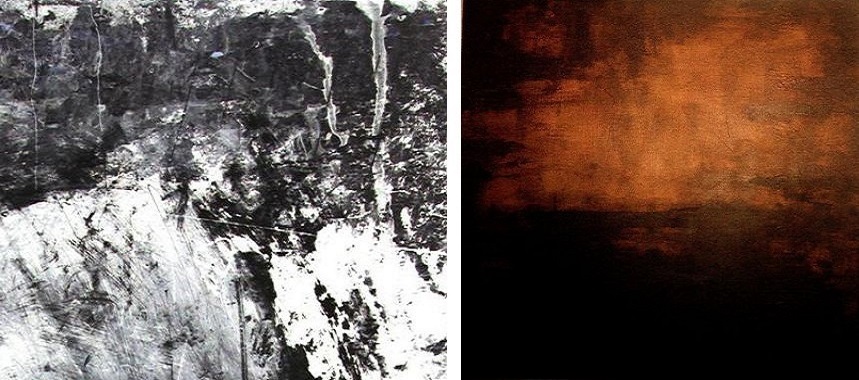}
}
\subfigure[Excitement]{
\includegraphics[width=0.4\textwidth]{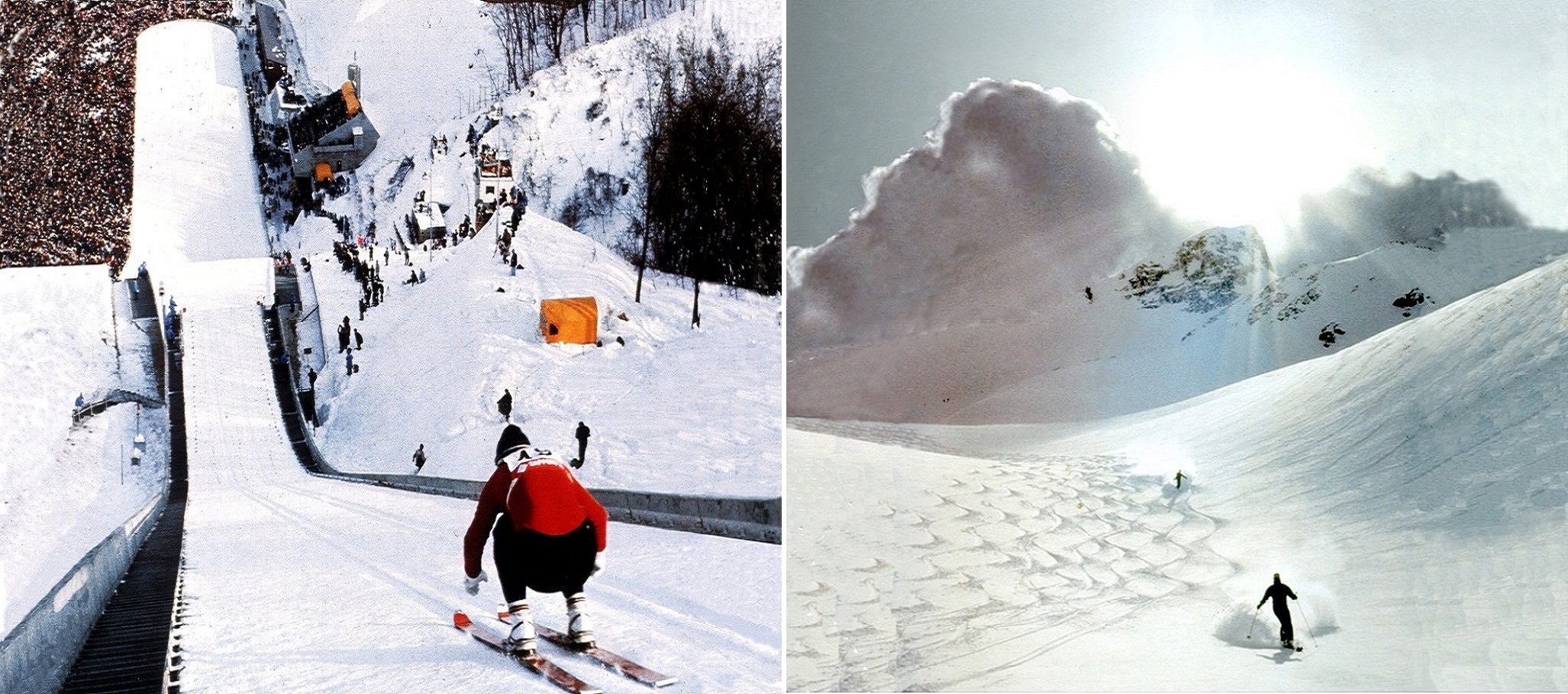}
}
\subfigure[Sadness]{
\includegraphics[width=0.4\textwidth]{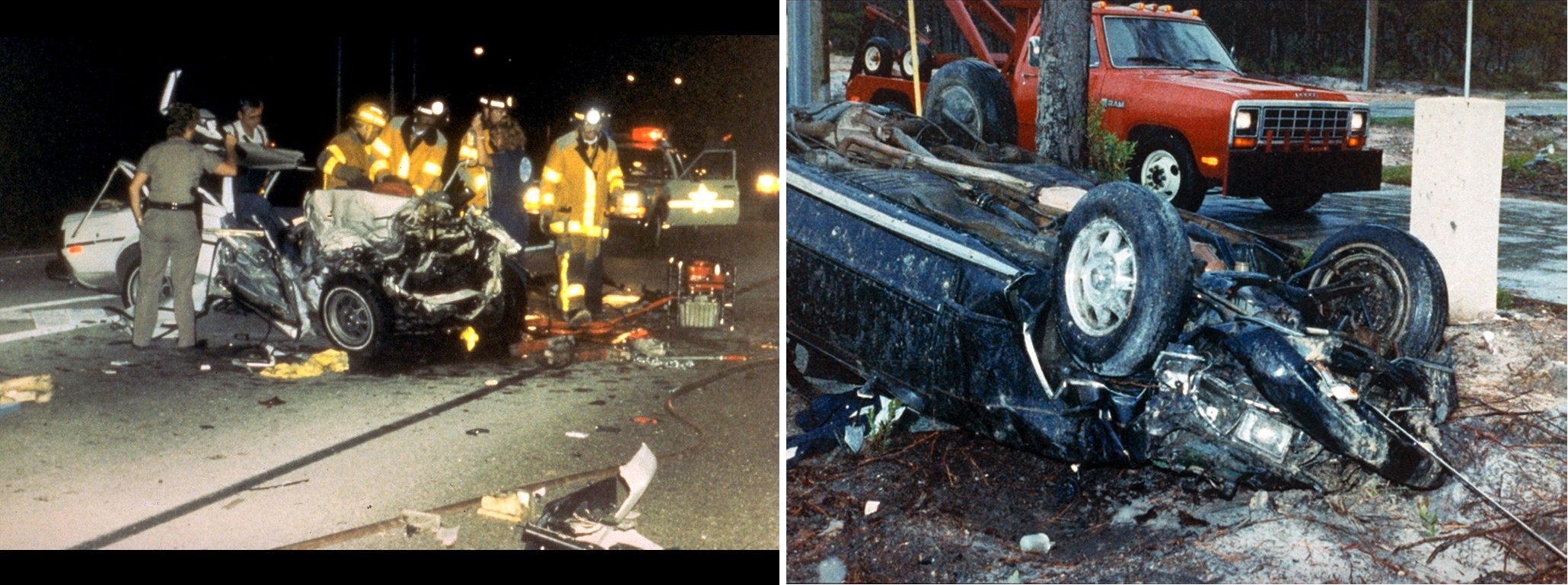}
}
\subfigure[Contentment]{
\includegraphics[width=0.4\textwidth]{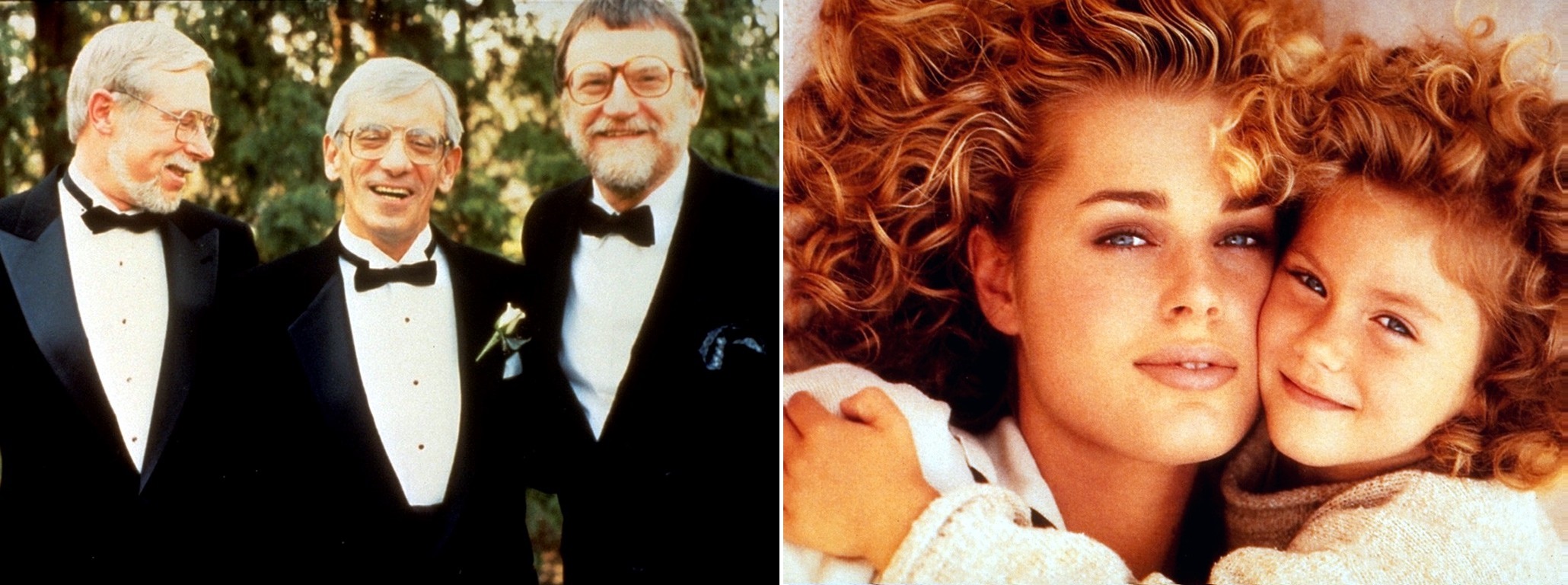}
}
\caption{The emotions conveyed by different kinds of images are correlated with different features~\cite{zhao2014affective}: (a) Aesthetic features (low saturation, cool color, low color difference); (b) Attributes (snow, skiing); (c) Semantic concepts described by adjective noun pairs (broken car); (d) Facial expressions (happiness).}
\label{fig:Examples}
\end{center}
\end{figure}

With the rapid development and popularity of social networks, such as Twitter\footnote{\url{https://twitter.com}}
and Sina Weibo\footnote{\url{http://www.weibo.com}}, people tend to express and share their opinions and emotions online using text, images, and videos. Understanding the information contained in the increasing repository of data is of vital importance to behavior sciences~\cite{pang2008opinion}, which aim to predict human decision making and enable wide applications, such as mental health evaluation~\cite{guntuku2019twitter}, business recommendation~\cite{pan2014travel}, opinion mining~\cite{tumasjan2010predicting}, and entertainment assistance~\cite{zhao2020emotion}.

Analyzing media data on an affective (emotional) level belongs to affective computing, which is defined as ``\textit{the computing that relates to, arises from, or influences emotions}"~\cite{picard2000affective}. The importance of emotions has been emphasized for decades since Minsky introduced the relationship between intelligence and emotion~\cite{minsky1988society}. One famous claim is ``\textit{The question is not whether intelligent machines can have any emotions, but whether machines can be intelligent without emotions.}'' Based on the types of media data, the research on affective computing can be classified into different categories, such as text~\cite{giachanou2016like,zhang2018deep}, image~\cite{zhao2018affective}, speech~\cite{schuller2018speech}, music~\cite{yang2012machine}, facial expression~\cite{li2020deep}, video~\cite{wang2015video,zhao2020endtoend}, physiological signals~\cite{alarcao2017emotions}, and multi-modal data~\cite{soleymani2017survey,poria2017review,zhao2019affective}. 

The adage ``\textit{a picture is worth a thousand words}'' indicates that images can convey rich semantics. Therefore, images are used as an important channel to express emotions. Image emotion analysis (IEA) has recently been paid much attention. As compared to analyzing the images' cognitive aspect that is related with objective content~\cite{hanjalic2006extracting}, such as object classification and semantic segmentation, IEA focuses on understanding what emotions can be induced by the images in viewers. The challenges of affective gap and perception subjectivity~\cite{zhao2018affective} make IEA a difficult task.

In this chapter, we concentrate on introducing recent advances on IEA -- especially our recent efforts from a computational perspective and on suggesting future research directions. First, we briefly introduce some popular emotion representation models from psychology in Sec.~\ref{sec:EmotionModels}, define corresponding key computational problems, and provide some representative supervised frameworks in Sec.~\ref{Sec:KeyProblems}. Second, we introduce the major challenges in IEA in Sec.~\ref{Sec:Challenges}. Third, we present some representative methods on different computational components, such as emotion feature extraction in Sec.~\ref{Sec:Features} and supervised classifier learning as well as domain adaptation in Sec.~\ref{Sec:LearningMethods}.
Then, we introduce some typical datasets for IEA evaluation in Sec.~\ref{Sec:Datasets} and investigate the performances of different features and classifiers on these datasets in Sec.~\ref{Sec:Experiment}, as emotions can be conveyed by various features, as shown in Figure~\ref{fig:Examples}. Finally, we give a discussion on what questions are still open and provide some suggestions for future research in Sec.~\ref{Sec:Conclusion}.







\section{Emotion Representation Models from Psychology}
\label{sec:EmotionModels}

Psychologists have proposed different theories to explain the what, how, and why behind human emotions~\cite{plutchik2013theories}. For example, the James-Lange theory suggests that emotions occur as a result of physiological reactions to events; the Cognitive Appraisal theory claims that the sequence of events first involves a stimulus, followed by thought, which then leads to the simultaneous physiological response and emotion.
Some other emotion theories include the Evolutionary theory, the Cannon-Bard theory, the Schachter-Singer Theory, and the Facial-Feedback theory~\cite{plutchik2013theories}.

Besides emotion, several other concepts (e.\,g., affect, sentiment, feeling, and mood) are also widely used in psychology. The difference or correlation of these concepts can be found in~\cite{munezero2014they}. In this chapter, we focus on a computational perspective and do not distinguish them clearly, except sentiment for positive/negative/neutral categories and emotion for more fine-grained definitions. Another relevant concept is about expected, induced, or perceived emotion. Expected emotion is the emotion that the image creator intends to make people feel, perceived emotion is what people perceive as being expressed, while induced/felt emotion is the actual emotion that is felt by a viewer. Interested readers can refer to~\cite{juslin2004expression} for more details. Unless otherwise specified, the emotion focused in this chapter is about induced emotion because of the dataset construction process.

To quantitatively measure emotion, psychologists have mainly employed two types of emotion representation models, categorical emotion states (CES) and dimensional emotion space (DES)~\cite{zhao2018affective}. For CES, a set of pre-selected categories is used to define emotions. Some popular CES models include binary sentiment (positive and negative, sometimes including neutral), Ekman's six basic emotions (happiness, surprise and \textit{negative} anger, disgust, fear, and sadness)~\cite{ekman1992argument}, and Mikels's eight emotions (amusement, anger, awe, contentment, disgust, excitement, fear, and sadness)~\cite{mikels2005emotional}. More diverse and fine-grained emotion categories are being increasingly considered. In Plutchik's emotion model~\cite{plutchik1980emotion}, each basic emotion category (anger, anticipation, disgust, fear, joy, sadness, surprise, and trust) is organized into three intensities. For example, the three intensities from low to high for surprise are distraction$\longrightarrow$surprise$\longrightarrow$amazement. Parrott represents emotions with a three-level hierarchy, i.\,e., primary (positive and negative), secondary (anger, fear, joy,
love, sadness, and surprise), and tertiary (25 fine-grained categories)~\cite{parrott2001emotions}. For DES, a 2D, 3D, or higher dimensional Cartesian space is employed to represent emotions, such as valence-arousal-dominance (VAD)~\cite{schlosberg1954three} and activity-temperature-weight~\cite{lee2011fuzzy}. VAD is the most widely used DES model, where `V' represents the pleasantness ranging from positive to negative, `A' represents the intensity of emotion ranging from excited to calm, and `D' represents the degree of control ranging from controlled to in control.

Intuitively, CES models are easy for users to understand, but limited emotion categories cannot well reflect the complexity and subtlety of emotions. Further, psychologists have not reached a consensus on how many categories should be included. Theoretically, all emotions can be measured as different coordinate points in the continuous Cartesian space. However, such absolute continuous values are difficult for non-experts to understand. Specifically, CES can be transformed to DES but not all Cartesian points can correspond to detailed categories~\cite{alarcao2018identifying}. For example, fear is often related to negative valence, high arousal, and low dominance. In this chapter, the employed CES models mainly include binary sentiment and Mikels's eight emotions, and VAD is employed as the DES model.


\section{Key Computational Problems \& Supervised Frameworks}
\label{Sec:KeyProblems}
Based on different emotion representation models, we can perform different IEA tasks: classification/retrieval based on CES, and regression/retrieval based on DES. Current methods mainly employ supervised methods with the help of available labeled datasets. In this section, we will define the key computational problems and provide representative supervised frameworks.

\subsection{Emotion Classification and Regression}
\label{Sec:Classification}
Suppose all images in the dataset are grouped into $K$ emotion categories, then emotion prediction can be conceived as a multi-class classification problem. Based on the model trained on given training samples, an emotion category that is most likely evoked in humans is assigned to a test image. Suppose we have $N$ training images $\{(\mathbf{x}_{i},y_{i})_{i=1}^N\}$, where $y_{i}\in{\{1,2,\cdots,K\}}$. Let $g_\mu(\mathbf{x})$ denote the feature extractor of image $\mathbf{x}$, and then our goal is to learn some model $h_\theta (g_\mu(\mathbf{x})):g_\mu(\mathbf{x})\rightarrow y$ that maps image features $g_\mu(\mathbf{x})$ to emotion labels $y$, where $\mu$ and $\theta$ are parameters. Usually, the learning process is transformed to a parameter optimization problem, which can be defined as 
\begin{equation}
\begin{aligned}
&J(\omega,\theta,\mu)=\sum_{i=1}^{N}f_\omega(h_\theta (g_\mu(\mathbf{x}_{i}),y_{i}),\\
&[\omega^{*},\theta^{*},\mu^{*}]=\text{arg}\mathop{\argmin}_{\omega,\theta,\mu}J(\omega,\theta,\mu),
\end{aligned}
\end{equation}
where $f_\omega(.,.)$ is a function with parameters $\omega$ to compute the loss function $J(\omega,\theta,\mu)$ between the predicted labels and the ground truth, and $\text{arg}\argmin$ is the argument of the minimum. Once we work out $\mu$ and $\theta$, given a test image $\mathbf{x}_{te}$, we can obtain the prediction label $h_\theta (g_\mu(\mathbf{x}_{te}))$.

Emotion regression assumes that emotions are represented by continuous dimensional values instead of discrete emotion labels, i.\,e., $y$ is continuous. Except this, the learning process of emotion regression is analogous to emotion classification.

\begin{figure*}[t]
\begin{center}
\centering \includegraphics[width=1.0\linewidth]{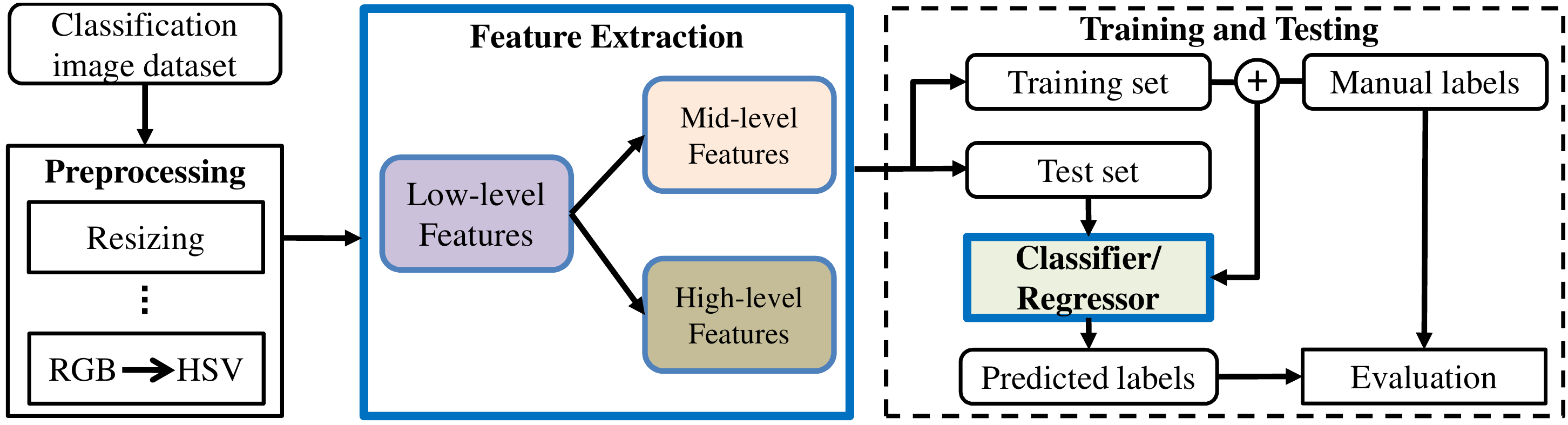}
\caption{Commonly used supervised framework of affective image classification and regression. The key components researchers have been studying lie in the solid blue rectangles.}
\label{fig:ClassificationFramework}
\end{center}
\end{figure*}

The commonly used supervised framework of affective image classification and regression is shown in Fig.~\ref{fig:ClassificationFramework}. Firstly, some preprocessing is done to `normalize' the images. Then,  different features are extracted for each image, which presents the core of image emotion analysis and will be described in detail. The dataset is split into a training set and a test set. A classifier or regressor is trained using the training set along with the emotion labels based on certain learning models. The images in the test set are then automatically classified by the trained classifier or regressed by the trained regressor. The assigned emotion labels are compared with the ground truth to evaluate the classification or regression performance.

\subsection{Emotion Retrieval}
\label{Sec:Retrieval}
Affective image retrieval, firstly named emotional semantic image retrieval~\cite{wang2008survey}, involves searching for images that express similar emotions to the query image. Affective image retrieval can be formalized as a reranking problem to ensure that the top ranked images are the ones emotionally similar to the query image.

\begin{figure}[t]
\begin{center}
\centering \includegraphics[width=0.8\linewidth]{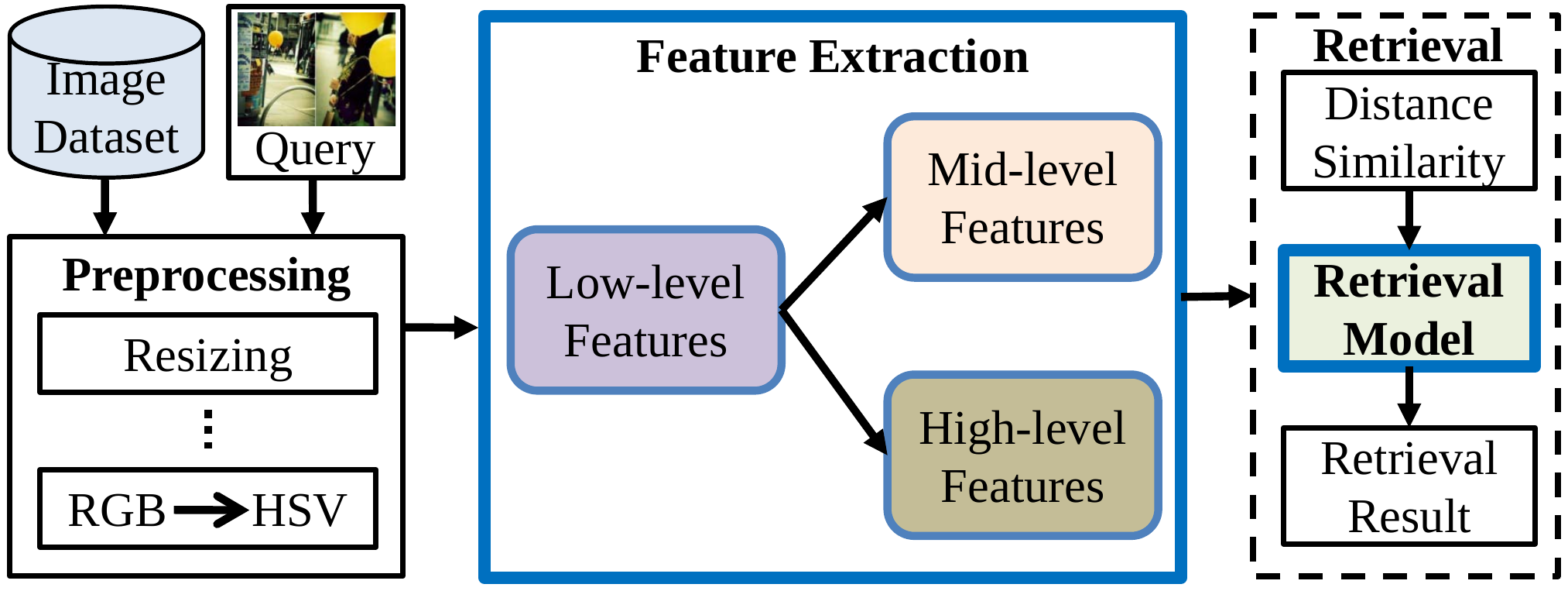}
\caption{Commonly used supervised framework of affective image retrieval. The key components researchers have been studying lie in the solid blue rectangles.}
\label{fig:RetrievalFramework}
\end{center}
\end{figure}

Suppose the features and emotion label of a given query image $\mathbf{x}_q$ are $g_\mu(\mathbf{x}_q)$ and $y_q$, and in the dataset there are $N_s$ emotionally similar images, in which the features and labels of the $i$th image are $\mathbf{x}_i^s$ and $y_i^s$, where $y_i^s==y_q, i=1,2,\cdots,N_s$, and $N_d$ emotionally different images, in which the features and labels of the $j$th image are $\mathbf{x}_j^d$ and $y_j^d$, where $y_j^d\not=y_q, j=1,2,\cdots,N_d$. Then, our goal is to minimize the distance between the query image and the $N_s$ positive images and maximize the distance between the query image and the $N_d$ negative images:
\begin{equation}
\small
\begin{aligned}
&J_s(\theta,\mu)=\sum_{i=1}^{N_s}h_\theta(D(g_\mu(\mathbf{x}_i^s),g_\mu(\mathbf{x}_q)),\\
&J_d(\theta,\mu)=\sum_{j=1}^{N_d}h_\theta(D(g_\mu(\mathbf{x}_j^d),g_\mu(\mathbf{x}_q)),\\
&J(\omega,\theta,\mu)=f_\omega(J_s(\theta,\mu),J_d(\theta,\mu)),\\
&[\omega^{*},\theta^{*},\mu^{*}]=\text{arg}\mathop{\argmin}_{\omega,\theta,\mu}J(\omega,\theta,\mu),
\end{aligned}
\label{Equ:retrievalFormulation}
\end{equation}
where $D(.,.)$ is a distance function to compute the distance between two feature vectors, such as the Minkowski-form distance and the Mahalanobis distance, $h_\theta(.)$  is a function with parameters $\theta$ to compute a cost of the query image and the image in the dataset, $f_\omega(.,.)$ is a function with parameters $\omega$ to compute the total cost $J(\omega,\theta,\mu)$ between the positive cost $J_s(\theta,\mu)$ and the negative cost $J_d(\theta,\mu)$. Once we work out $\mu$ and $\theta$, we can get the retrieval results by sorting the cost.

The commonly used supervised framework of affective image retrieval is shown in Fig.~\ref{fig:RetrievalFramework}. The preprocessing and feature extraction parts are similar to the related parts in emotion classification and regression. The distance or similarity is computed between the features of the query image and each image in the dataset. Through some retrieval model, we sort the distance or similarity and obtain the retrieval results, which are compared with the ground truth for evaluation.

\section{Major Challenges}
\label{Sec:Challenges}


\textbf{Affective Gap.}
The affective gap is one main challenge for IEA, which is defined as the inconsistency between extracted low-level features and induced emotions~\cite{hanjalic2006extracting,zhao2018affective}. As compared to the semantic gap in computer vision, i.\,e., the discrepancy between the limited descriptive power of low-level visual features and the richness of user semantics~\cite{smeulders2000content,liu2007survey}, the affective gap is even more challenging. Bridging the semantic gap cannot guarantee bridging the affective gap. For example, images containing a barking dog and a loving dog are both about dogs but obviously induce different emotions. To bridge the affective gap, the main efforts have been focusing on designing and extracting discriminative emotion features, ranging from the early hand-crafted features to more recent deep ones. Based on these features, a dominant emotion category (DEC) is assigned to an image by traditional single-label learning-based methods.

\textbf{Perception Subjectivity.}
Emotion is a highly subjective and complex variable. Different viewers may perceive totally different emotions to the same image, which is influenced by many factors, such as culture, education, personality, and environment~\cite{zhao2018predicting}. For example, for a sudden heavy snow, some may feel excitement to see such rare natural scenes, some may feel sadness because the planned activities have to be cancelled, some may feel amusement since they can build a snowman, etc. For the subjectivity challenge, one direct and intuitive solution is to predict  emotions for each viewer via personalized learning models~\cite{zhao2018predicting}. When a large number of viewers are involved, we can assign the image with multiple emotion labels via multi-label learning methods. Since the importance or extent of different labels is actually unequal, predicting the probability distribution of emotions, either discrete~\cite{yang2017joint,zhao2020discrete} or continuous~\cite{zhao2017continuous}, would make more sense.

\textbf{Label Noise and Absence.}
Recent deep learning based IEA methods have achieved state-of-the-art performances with the help of large-scale labeled training data. However, in real applications, it is expensive and time-consuming and even impossible to obtain sufficient data with emotion labels to train a deep model. It would be more practical if we can deal with the situation that there are only few or even no emotion labels. We can conduct unsupervised/weakly supervised learning~\cite{wei2020learning} and few/zero shot learning~\cite{zhan2019zero}. One might consider leveraging the large amount of weakly-labeled web images~\cite{wei2020learning}. Since the associated tags might contain noise that is unrelated to emotion and even to visual semantics, filtering such automatic labels is necessary. Another possible solution is to transfer the well-learned model on one labeled source domain to another unlabeled or sparsely labeled target domain. Direct transfer often results in obvious performance decay, because of the influence of domain shift~\cite{zhao2020review}, i.\,e., the joint distribution of images and emotion labels are different across domains. To bridge the domain shift challenge, we can employ domain adaptation and domain generalization techniques~\cite{zhao2021emotional}.

\section{Emotion Features}
\label{Sec:Features}
In this section, we summarize the features that have been widely extracted for IEA, including both hand-crafted and deep features. We first give an brief overview and then introduce some representative ones especially our recent work.

\subsection{Hand-crafted Features}
\label{ssec:Hand-crafted}

\textbf{Overview.} Early efforts on IEA mainly focused on hand-crafting features from different levels. \textit{Low-level features} are used in the earliest IEA methods, which suffer from large affective gap and low interpretability. Some generic features from computer vision, such as Gabor, HOG, and GIST, are directly used in the IEA task~\cite{yanulevskaya2008emotional}. Some specific features derived from elements of art, including color and texture, are implemented~\cite{machajdik2010affective}. Low-level color features include mean saturation and brightness, vector based mean hue, emotional coordinates (pleasure, arousal and dominance) based on brightness and saturation, colorfulness and color names. Low-level texture features include Tamura texture, Wavelet textures, and gray-level co-occurrence matrix (GLCM) based texture~\cite{machajdik2010affective}. Low-level shape features, including line segments, angles, continuous lines, and curves, are designed in~\cite{lu2012shape}. As compared to low-level features, \textit{mid-level features} are more interpretable, semantic, and relevant to emotions. Different types of attributes people use to describe scenes, such as materials, surface properties, functions or affordances, spatial envelope attributes, and object presence are modeled~\cite{yuan2013sentribute}. Features inspired from principles of art, such as symmetry, emphasis, harmony, and variety, are specially designed~\cite{zhao2014exploring}. \textit{High-level features} describe the detailed content in an image through which viewers can easily understand the semantics and evoked emotions. Some representative high-level features include adjective noun pairs detected by SentiBank~\cite{borth2013large} and recognized facial expressions~\cite{yang2010exploring}.

\begin{figure*}[t]
\begin{center}
\centering \includegraphics[width=1.0\linewidth]{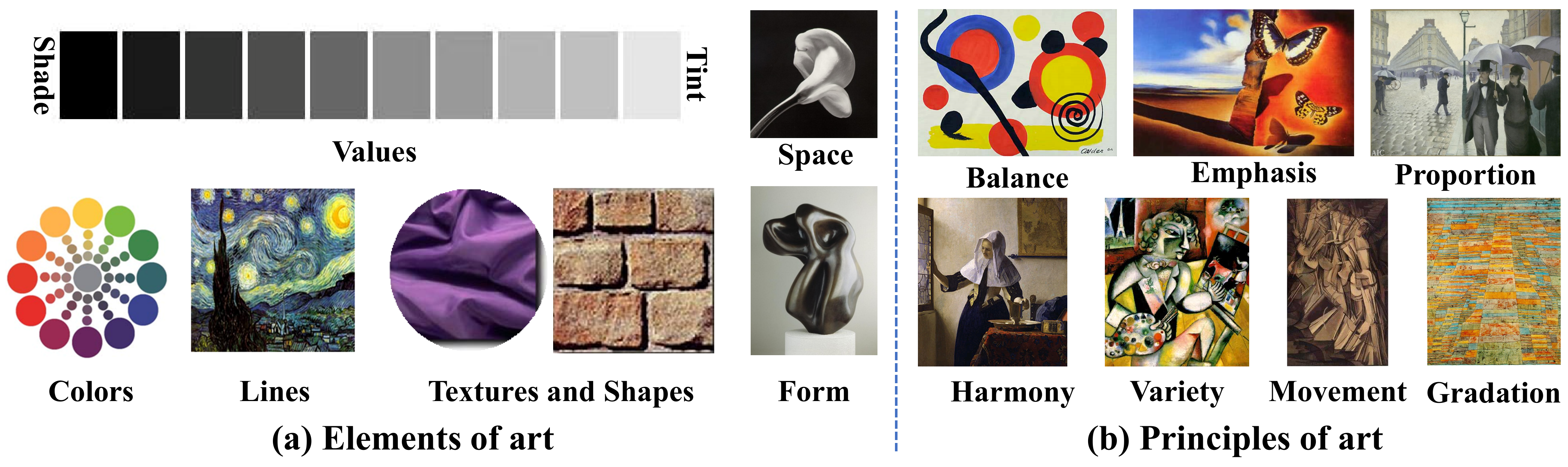}
\caption{Illustration of artistic elements and artistic principles, which are designed as low-level and mid-level emotion features.}
\label{fig:PAEF}
\end{center}
\end{figure*}

\textbf{Mid-level Principles-of-art Based Emotion Features.} The principles of art are defined as the rules, tools, or guidelines of arranging and orchestrating the elements of art in an artwork. They consider various artistic aspects including balance, emphasis, harmony, variety, gradation, movement, rhythm, and proportion~\cite{zhao2014exploring}. The comparison of elements of art and principles of art is shown in Fig.~\ref{fig:PAEF}. Six principles of art are formulated and implemented systematically in~\cite{zhao2014exploring} based on related art theory and multimedia research. Totally, a 165 dimensional feature can be obtained for each image. For example, emphasis, also known as contrast, is used to stress the difference of certain elements, which can be accomplished by using sudden and abrupt changes in elements. Itten color contrast, which is defined to coordinate colors using the hue's contrasting properties, is implemented~\cite{zhao2014exploring}, including contrast of saturation, contrast of light and dark, contrast of extension, contrast of complements, contrast of hue, contrast of warm and cold, and simultaneous contrast. The results show that principles of art features are more correlated with emotions than elements of art~\cite{zhao2014exploring}. For example, images with high balance and harmony values tend to express positive emotions.

\textbf{High-level Adjective Noun Pairs.} The adjective noun pairs (ANPs) are detected by a large detector library SentiBank~\cite{borth2013large}, which is trained using GIST, a $3\times 256$ dimension color histogram, a 53 dimensional LBP descriptor, a Bag-of-Words quantized descriptor using a 1,000 word dictionary with a 2-layer spatial pyramid and max pooling, and a 2,000 dimensional attribute on about 500k images downloaded from Flickr. Liblinear support vector machine (SVM)~\cite{fan2008liblinear} is used as classifier and early fusion is adopted. The advantages of ANP are that it turns a neutral noun into an ANP with strong emotions and makes the concepts more detectable, as compared to nouns and adjectives, respectively. Finally, a 1,200 dimensional double vector representing the probability of the ANPs is obtained.

\subsection{Deep Features}
\label{ssec:DeepFeatures}

\textbf{Overview.} With the development of deep learning, especially convolutional neural networks (CNNs), learning-based deep features have been widely employed with superior performances as compared to hand-crafted ones. \textit{Global features} are directly extracted from the whole images. One direct and intuitive method is to employ the output of the last few fully connected (FC) layers as deep features, using either pretrained or finetuned CNN models~\cite{xu2014visual,chen2015learning,you2016building}. The last few FC layers correspond to high-level semantic features, which might be not enough to represent emotions, especially for abstract images. Therefore, some methods try to extract multi-level deep features~\cite{rao2020learning,zhu2017dependency,yang2018retrieving}. For example, three parallel networks, namely an Alexnet, an aesthetics CNN, and a texture CNN, are trained with different levels of image patches as input. Deep representations at three levels, i.\,e., image semantics, image aesthetics, and low-level visual features are extracted. The features from different layers in CNNs are extracted as multi-level representations, which are fed into a bidirectional gated recurrent unit model to exploit the dependency among different levels of features~\cite{zhu2017dependency}. The above methods treat different regions of an image equally. Based on the fact that some regions can determine the emotion of an image while the other regions do not help much and might even reverse, some recent methods focus on extracting \textit{local features} that are more discriminative for IEA~\cite{you2017visual,she2020wscnet,zhao2019pdanet,yao2020apse}.

\textbf{Weakly Supervised Coupled Networks (WSCNet).} WSCNet contains two branches for joint emotion detection and classification~\cite{she2020wscnet}. One is the detection branch which is designed to generate region proposals that evoke emotion. A soft sentiment map is generated by a cross-spatial pooling strategy to summarize all the information contained
in the feature maps for each category. The regions of interest that are informative for classification are highlighted in the sentiment map. The advantage of such setting is that the network can be trained with image-level emotion labels, without requiring time-consuming region-level annotation. The other is the classification branch designed for the emotion classification task by considering both global and local representations. The global features are extracted from a fully convolutional network (FCN), while the local features are obtained by coupling the generated sentiment map in the detection branch with the global features.

\begin{figure}[t]
\begin{center}
\centering \includegraphics[width=1.0\linewidth]{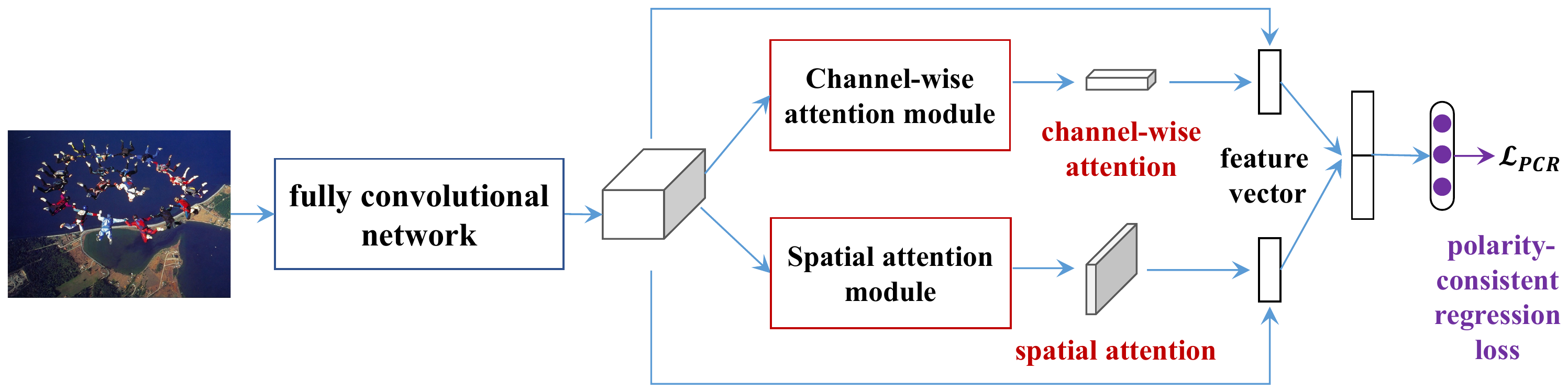}
\caption{Overview of the polarity-consistent deep attention network (PDANet)~\cite{zhao2019pdanet} to extract attended features for IEA.}
\label{fig:PDANet}
\end{center}
\end{figure}

\textbf{Polarity-consistent Deep Attention Network (PDANet).} The feature maps of PDANet from a FCN are fed into two branches~\cite{zhao2019pdanet}, as shown in Fig.~\ref{fig:PDANet}. Each branch is a multi-layer neural network. One is used to estimate the spatial attention to emphasize the emotional semantic-related regions by two $1\times 1$ convolutional layers and a hyperbolic tangent function. The other is used to estimate the channel-wise attention to consider the interdependency between different channels by one $1\times 1$ convolutional layer and a sigmoid function. The attended semantic vectors that capture the global and local information respectively are concatenated as the final feature representations for IEA tasks.

\textbf{Attention-aware Polarity-Sensitive Embedding (APSE).} APSE utilizes a hierarchical attention mechanism to learn both polarity and emotion-specific attended representations~\cite{yao2020apse}. Based on the fact that concrete emotion categories depend on high-level semantic information and that polarity is relevant to low-level features (e.\,g., color and texture), polarity-specific attention is modeled in lower layers and emotion-specific attention is modeled in higher layers. These two types of attended features are integrated by cross-level bilinear pooling to facilitate the interaction between the information of different levels. After  dimensionality reduction and $\ell_2$-Normalization, we can obtain the final feature representations.

\section{Learning Methods for IEA}
\label{Sec:LearningMethods}
In this section, we first summarize the supervised learning methods that have been widely used for emotion classification, regression and retrieval. Then, we introduce some domain adaptation methods.

\subsection{Emotion Classification}
\label{sec:LearningClassification}

\textbf{Shallow Pipeline.} 
Based on the modeling process, supervised learning can be classified into generative learning and discriminative learning. Discriminative learning models the conditional distribution of labels $y$ given features $g_\mu(\mathbf{x})$ directly or learns the mappings directly from features $g_\mu(\mathbf{x})$ to labels $y$. For instance, logistic regression, a binary classification method, models the conditional distribution $p(y|g_\mu(\mathbf{x});\theta)$ as:
\begin{equation}
h_\theta(g_\mu(\mathbf{x}))=\text{sig}(\theta^{T}g_\mu(\mathbf{x})),    
\end{equation}
where $\text{sig}$ is the sigmoid function $\displaystyle \text{sig}(z)=\frac{1}{1+e^{-z}}$ and $\theta$ is the vector of parameters. A generalization of logistic regression to multi-class classification is softmax regression. The perceptron learning algorithm `forces' the output values of logistic regression to be exactly 0 or 1, based on the threshold function:
\begin{equation}
\text{sig}(z)=\begin{cases}
1, & \text{if } z\ge 0,\\
0, & \text{if } z<0.
\end{cases}    
\end{equation}
Support vector machines (SVM) try to find a decision boundary that maximizes the geometric margin and can be extended with various non-linear kernels.

Generative learning algorithms try to model class priors $p(y)$ and likelihood $p(g_\mu(\mathbf{x})|y)$, and then, the posterior distribution on $p(y|g_\mu(\mathbf{x}))$ can be derived by Bayes rule:
\begin{equation}
\displaystyle p(y|g_\mu(\mathbf{x}))=\frac{p(g_\mu(\mathbf{x})|y)p(y)}{p(g_\mu(\mathbf{x}))},    
\end{equation}
where $p(g_\mu(\mathbf{x}))$ can be seen as a normalization factor. Gaussian discriminant analysis assumes that $p(g_\mu(\mathbf{x})|y)$ is distributed according to a multivariate Gaussian distribution, which deals with continuous real-valued features. Naive Bayes, which handles discrete values of $g_\mu(\mathbf{x})$, is based on the assumption that the discrete values are conditionally independent given $y$. When dealing with multi-class classification, it is often formulated as some extensions of binary classification. The prominent formulations include `one-versus-all' and `one-versus-one' classification.

\textbf{Deep Architecture.} Recent deep learning based emotion classification methods usually employ several fully-connected (FC) layers to minimize the following cross-entropy loss~\cite{she2020wscnet}:
\begin{equation}
{\mathcal{L}_{CE}} =  - \frac{1}{N}\sum\limits_{i = 1}^N {\sum\limits_{k = 1}^K {\mathds{1}_{[k=y_i]}\log p_{i, k}} },
\label{equ:ceLoss}
\end{equation}
where $K$ is the number of emotion classes, $\mathds{1}_{[k=y_i]}$ is a binary indicator, and $p_{i, k}$ is the predicted probability that image $i$ belongs to class $k$. Directly optimizing the cross-entropy loss might lead some images to be incorrectly classified into categories with opposite polarity. For example, for an image with the emotion ``amusement'', one model might classify the emotion incorrectly as ``sadness'' which has an opposite polarity (negative vs.\ positive). But it is more acceptable if the emotion is classified as ``excitement'' which has the same polarity (positive). Based on this motivation, a novel polarity-consistent cross-entropy (PCCE) loss is proposed to consider the polarity-emotion hierarchy by increasing the penalty of the predictions that have opposite polarity to the ground truth~\cite{zhao2020endtoend}. The PCCE loss is defined as:
\begin{equation}
{\mathcal{L}_{PCCE}} =  - \frac{1}{N}\sum\limits_{i = 1}^N  (1 + \lambda (G({\hat{y}_i, y_i}))) {\sum\limits_{k = 1}^K {\mathds{1}_{[k=y_i]}\log {p_{i,k}}} },
\label{equ:pcceLoss}
\end{equation}
where $\lambda$ is a penalty coefficient. Similar to the indicator function, $G(.)$ represents whether to add the penalty or not and is defined as:
\begin{equation}
G(\hat{y},y)=\begin{cases}
1, & \text{if } \text{polarity}(\hat{y})\neq \text{polarity}(y),\\
0, & \text{otherwise},
\end{cases}
\label{equ:indicator}
\end{equation}
where $\text{polarity}(.)$ is a function that maps an emotion category to its polarity (positive or negative).

\subsection{Emotion Regression}
\label{sec:LearningRegression}

In the early shallow pipeline, some commonly used regression methods, including linear regression, support vector regression (SVR), and manifold kernel regression, are employed to predict the average dimensional values. For example, SVR is used in~\cite{lu2012shape} to predict emotion scores in the VA space.

Similar to emotion classification, deep learning based emotion regression methods also employ several fully-connected (FC) layers
to minimize the following mean squared error (MSE):
\begin{equation}
\mathcal{L}_{reg}=\frac{1}{N}\sum_{i=1}^{N}\sum_{j=1}^{N_E}(\hat{y}_i^j-y_i^j)^2,
\label{equ:mseLoss}
\end{equation}
where $N_E$ is the dimension number of the adopted emotion model ($N_E=3$ for VAD), and $y_i^j$ indicates the emotion label of the $j$-th dimension for image $\mathbf{x}_i$. Similar to PCCE loss, polarity-consistent regression (PCR) loss is proposed based on the assumption that VAD dimensions can be classified into different polarities~\cite{zhao2019pdanet}. The PCR loss is defined as:
\begin{equation}
\mathcal{L}_{PCR}=\frac{1}{N}\sum_{i=1}^{N}\sum_{j=1}^{N_E}(\hat{y}_i^j-y_i^j)^2(1+\lambda G({\hat{y}_i^j, y_i^j})).
\label{equ:pcrLoss}
\end{equation}

\subsection{Emotion Retrieval}
\label{sec:LearningRetrieval}

We introduced our work on multi-graph learning (MGL)~\cite{zhao2014affective} and attention-aware polarity-sensitive embedding (APSE)~\cite{yao2020apse} as shallow and deep methods for emotion retrieval. As a (semi-)supervised learning, MGL is widely used for reranking in various domains. For each feature, we can construct a single graph, where the vertices represent image samples and the edges reflect the similarities between sample pairs. By combining the multiple graphs together in a regularization framework, we can learn the optimized weights of each graph to efficiently explore the complementarity of different features~\cite{zhao2014affective}.

Besides the polarity and emotion-specific attended representations, APSE also consists of a polarity-sensitive
emotion-pair (EP) loss to further exploit the polarity-emotion hierarchy~\cite{yao2020apse}. Suppose $K$ pairs of convolution features constructed from $K$ different categories are formulated as $\left\{(g_1,g^+_1),\cdots,(g_K,g^+_K)\right\}$, where $g_k$ and $g_k^+$ represent the feature representations of anchor point $\mathbf{x}_k$ and positive example $\mathbf{x}_k^+$, respectively, both from the $k^{\rm th}$ category. The EP loss is the combination of inter-polarity loss and intra-polarity loss. Specifically, inter-polarity loss is formulated as:
\begin{equation}
\label{equ:inter-polarity}
\mathcal{L}_{inter} = \frac{1}{K}\sum_{k=1}^{K}\log(1+\exp(\frac{1}{N_{\mathcal{Q}_k}}\sum_{j\in\mathcal{Q}_k} g_k^\top g_j^+
-\frac{1}{N_{\mathcal{P}_k}}\sum_{j\in\mathcal{P}_k,j\neq k} g_k^\top g_j^+)),
\end{equation}%
where $\mathcal{P}_k$ and $\mathcal{Q}_k$ represent the sets of emotion categories in the same and opposite polarities to the anchor of the $k^{\rm th}$ category, respectively.
$N_{\mathcal{P}_k}$ and $N_{\mathcal{Q}_k}$ are the numbers of corresponding categories. The intra-polarity loss that can differentiate similar categories within the same polarity is defined as: 
\begin{equation}
\label{equ:intra-polarity}
\mathcal{L}_{intra} = \frac{1}{K}\sum_{k=1}^{K}\log(1+\sum_{j\in\mathcal{P}_k,j\neq k}\exp(g_k^\top g_j^+-g_k^\top g_k^+)).
\end{equation}%

\subsection{Emotion Distribution Learning}
\label{sec:DistributionLearning}

Emotion distribution learning is essentially a regression problem. We can directly employ regression methods to predict the probabilities of each emotion category, but the relationship between different emotion categories is ignored. Shared sparse learning (SSL) is employed to learn the probabilities of different emotion categories simultaneously  as a distribution~\cite{zhao2020discrete}. SSL is performed based on two assumptions: (1) the images, which are close to one another in the visual feature space, would have similar emotion distributions in the categorical emotion space; (2) the distribution of a test image can be approximately modeled as a linear combination of the distributions of the training images. Specifically, the combination coefficients are learned in the feature space and transferred to the emotion distribution space. The method is also extended to a more general setting, where multiple features are available. The optimal weights for each feature are automatically learned to reflect the importance of different features.

One intuitive method using deep architecture is to replace the cross-entropy loss for classification with some distribution-based losses, such as KL divergence~\cite{yang2017joint}:
\begin{equation}
{\mathcal{L}_{KL}} =  - \frac{1}{N}\sum\limits_{i = 1}^N {\sum\limits_{k = 1}^K {y_i^j \ln \hat{y}_i^j} },
\label{equ:klloss}
\end{equation}
where $y_i^j$ and $\hat{y}_i^j$ are the ground truth and predicted probability of the $j$th emotion category for image $\mathbf{x}_i$. The joint classification and distribution learning (JCDL) models both emotion classification and distribution learning simultaneously~\cite{yang2017joint}.

\begin{figure}[!t]
\begin{center}
\includegraphics[width=0.98\linewidth]{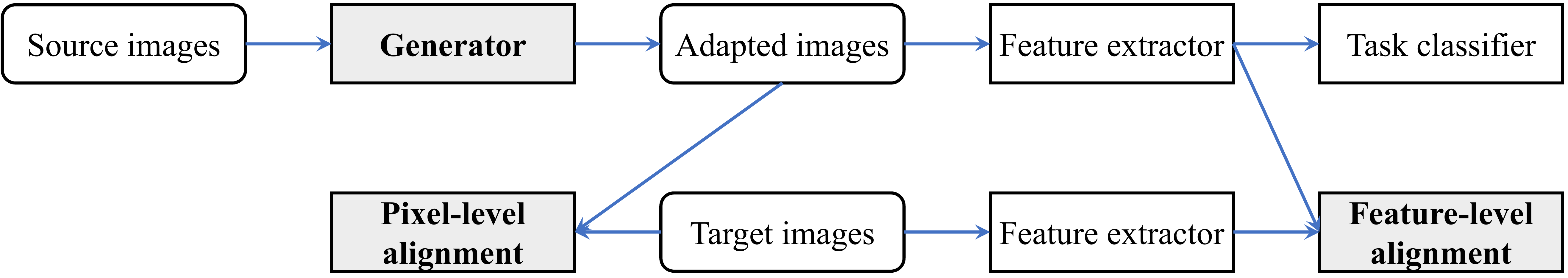}
\caption{A generalized domain adaptation framework for IEA with one labeled source domain and one unlabeled target domain. The gray-scale rectangles represent different alignment strategies. Most existing domain adaptation methods can be obtained by employing different component details, enforcing some constraints, or slightly changing the architecture.}
\label{fig:domainAdaptation}
\end{center}
\end{figure}

\subsection{Domain Adaptation}
\label{sec:DomainAdaptation}
Domain adaptation aims to learn a transferable model from a labeled source domain that can perform well on another sparsely labeled or unlabeled target domain~\cite{zhao2020review}. Most recent methods focused on the unsupervised setting with a two-stream deep architecture: one stream for training a task model on the labeled source domain, and the other stream for aligning the source and target domains, as shown in Fig.~\ref{fig:domainAdaptation}. The main difference of existing domain adaptation methods lies in the alignment strategy, which includes discrepancy-based, adversarial discriminative, adversarial generative, and self-supervision-based methods~\cite{zhao2020review}.

CycleEmotionGAN++ (CEGAN++)~\cite{zhao2021emotional} is one state-of-the-art domain adaptation method for IEA. CEGAN++ aligns the source and target domains at both pixel-level and feature-level. First, an  adapted domain is generated to perform pixel-level alignment by improving CycleGAN~\cite{zhu2017unpaired} with a multi-scale structured cycle-consistency loss. Dynamic emotional semantic consistency (DESC) is enforced to preserve the emotion labels of the source images during image translation. Second, feature-level alignment is conducted when learning the task classifier. The final objective loss is the combination of task loss, mixed CycleGAN loss, and DESC loss.

\section{Released Datasets}
\label{Sec:Datasets}
In this section, we introduce some datasets that are widely used for performance evaluation of IEA. For clarity, we organize these datasets based on different emotion labels and IEA tasks, i.\,e., average dimensional values, dominant emotion category, probability distribution, and personalized emotion labels.

\textbf{Average Dimensional Values.} \textit{The International Affective Picture System} (\textit{IAPS}) \cite{lang1997international} is an emotion evoking image set in psychology with 1,182 documentary-style natural color images depicting complex scenes, such as portraits, babies, animals, landscapes, etc. Each image is associated with an empirically derived mean and standard deviation (STD) of VAD ratings in a 9-point rating scale by about 100 college students (predominantly US-American). \textit{The Nencki Affective Picture System} (\textit{NAPS})~\cite{marchewka2014nencki} is composed of 1,356 realistic, high-quality photographs with five categories, i.\,e., people, faces, animals, objects, and landscapes.  204 mostly European participants labeled these images in a 9-point bipolar semantic sliding scale on the VA and approach-avoidance dimensions.
\textit{The Emotions in Context Database} (\textit{EMOTIC})~\cite{kosti2017emotion} consists of 18,316 images about people in context in non-controlled environments. There are two kinds of emotion labels: 26 emotion categories and the continuous 10-scale VAD dimensions.

\begin{figure}[tb]
\begin{center}
\subfigure[IAPS dataset]{
\includegraphics[width=0.48\textwidth]{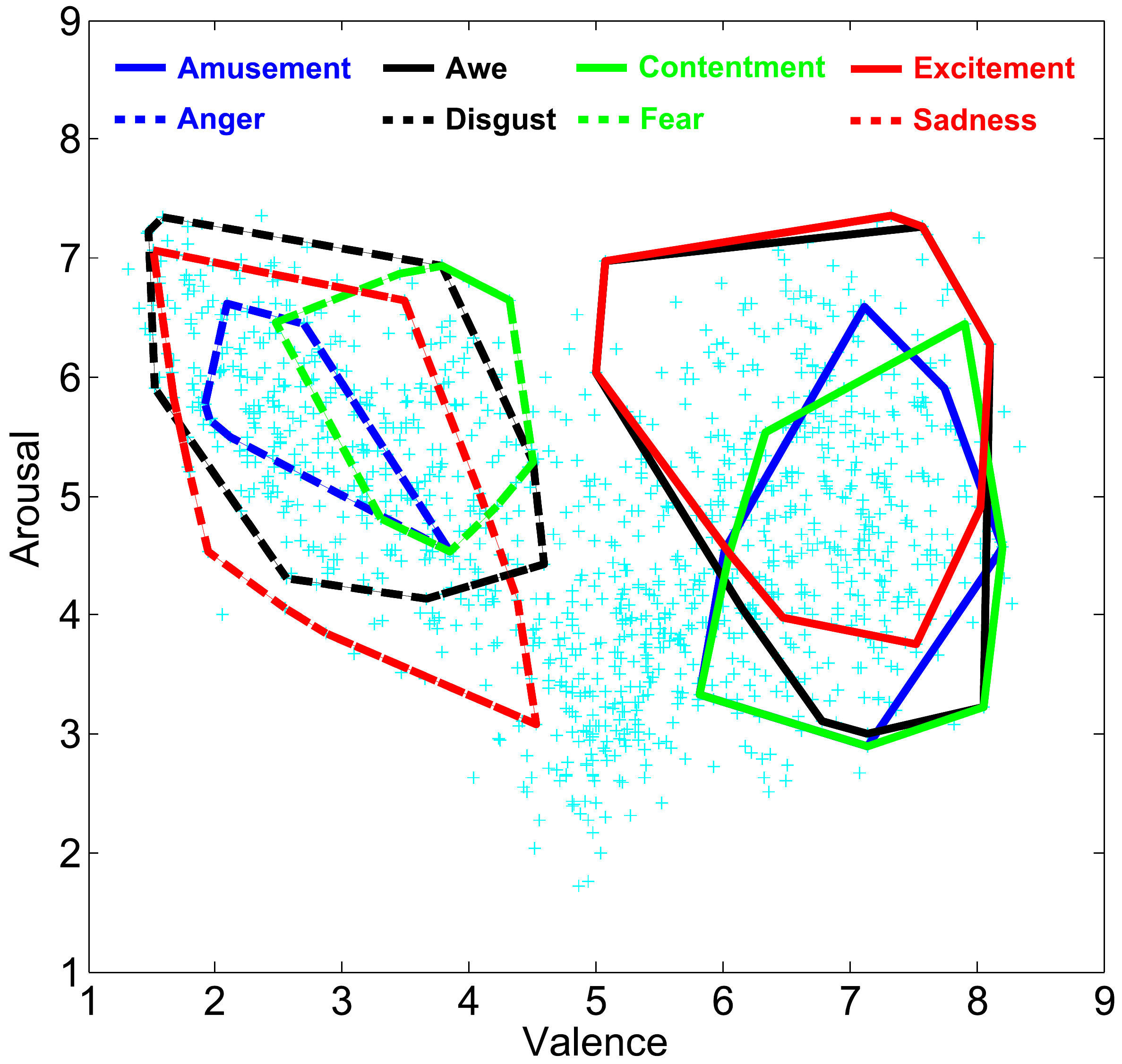}
\label{fig:IAPS}
}
\subfigure[GAPED dataset]{
\includegraphics[width=0.48\textwidth]{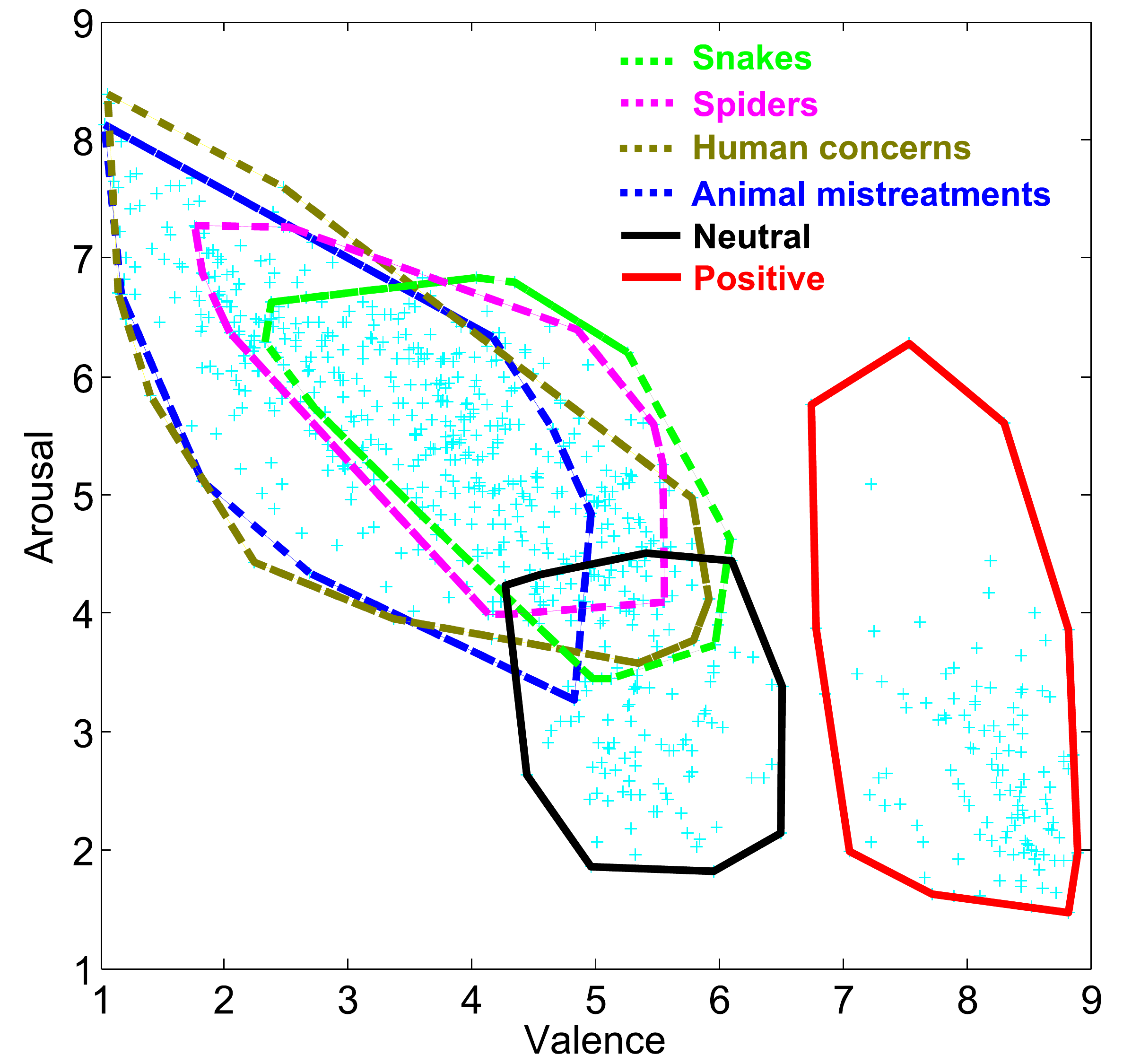}
\label{fig:GAPED}
}
\caption{Representation of the outcome ratings in the valence/arousal space of the IAPS and GAPED datasets. Polygons represent the surfaces occupied by all the images in a given category.}
\label{fig:Datasets}
\end{center}
\end{figure}

\textbf{Dominant Emotion Category.} \textit{IAPSa}~\cite{mikels2005emotional} is subset of IAPS, which includes 246 images. \textit{Abstract dataset} (\textit{Abstract}) contains 228 peer rated abstract paintings without contextual content~\cite{machajdik2010affective}. \textit{ArtPhoto} is an artistic dataset with 806 art photos obtained from a photo sharing site~\cite{machajdik2010affective}.
The \textit{IAPSa}, \textit{Abstract}, and \textit{ArtPhoto} datasets are categorized into eight discrete categories~\cite{mikels2005emotional}: amusement, anger, awe, contentment, disgust, excitement, fear, and sadness. The relationship between emotion categories and dimensional VA values is summarized in Fig.~\ref{fig:IAPS}. \textit{The Geneva affective picture database} (\textit{GAPED}) consists of 520 negative (133 spiders, 158 snakes, 105 human concerns, and 124 animal mistreatment) images, 121 positive (human and animal babies and nature sceneries) images and 89 neutral (inanimate objects) images~\cite{dan2011geneva}. Besides, these images are also rated with valence and arousal values, ranging from 0 to 100 points. The valence and arousal ratings (changed from $[0,100]$ to $[1,9]$) are shown in Fig.~\ref{fig:GAPED}.
\textit{Twitter I}~\cite{you2015robust} consists of 1,269 images annotated by 5 Amazon Mechanical Turk (AMT) workers. There are three subsets, i.\,e., ``Five agree'' (Twitter I-5), ``At least four agree'' (Twitter I-4), and ``At least three agree'' (Twitter I-3). ``Five agree'' indicates that all the 5 AMT workers labeled the same sentiment label to an image. There are 882 ``Five agree'' images and all the images receive at least three same votes. \textit{Twitter II} includes 470 positive tweets and 133 negative tweets~\cite{borth2013large} crawled from PeopleBrowsr with 21 hashtags. \textit{EMOd}~\cite{fan2018emotional} consists of 1,019 emotional images with eye-tracking data and different kinds of labels, such as object contour and emotions.
\textit{FI}~\cite{you2016building} is a large-scale  image emotion dataset with 23,308 images labeled using Mikel's emotion categories. The images are obtained by searching from Flickr and Instagram with the eight emotions as keywords and removing noisy data.

%


\textbf{Probability Distribution.}
The \textit{Flickr\_LDL} and \textit{Twitter\_LDL} datasets are constructed to study emotion ambiguity~\cite{yang2017learning}. There are 10,700 images and 10,045 images in these two datasets, which are labeled by 11 and 8 participants based on Mikel's emotion categories, respectively. Based on the detailed annotations, we can easily obtain the discrete probability distribution of different emotion categories.

\textbf{Personalized Emotion Labels.}
\textit{Image-Emotion-Social-Net} (\textit{IESN})~\cite{zhao2018predicting} is constructed to study personalized emotions. There are more than one million images crawled from Flickr uploaded by 11,347 users. For each image, both the expected emotion from the uploader and actual emotion from each viewer are provided in terms of binary sentiment, Mikel's emotion categories, and continuous VAD values.

\section{Experimental Results and Analysis}
\label{Sec:Experiment}

To give readers a clear understanding of the capabilities of current computational IEA methods, we conduct a series of experiments on different IEA tasks. In this section, we first introduce the evaluation criteria and then report the performance comparison of different representative methods.

\subsection{Evaluation Criteria}
\label{sec:EvaluationCriteria}

For emotion classification, the most widely used metric is classification accuracy, which measures the percentage of correctly classified images over all test images~\cite{she2020wscnet}. For emotion regression, we can use mean squared error, mean absolute error, and the coefficient of determination to evaluate the results~\cite{zhao2019pdanet}. For emotion distribution learning, we can either use the sum of squared difference to measure the performance from
the aspect of regression~\cite{zhao2020discrete}, or use distance or similarity metrics (e.\,g., KL divergence, Bhattacharyya coefficient, Chebyshev distance, Clark distance, Canberra metric, cosine coefficient, and intersection similarity) between two distributions to measure whether the predicted distribution and the ground truth is similar~\cite{yang2017joint,zhao2020discrete}. For image retrieval, there are several evaluation metrics: nearest neighbor rate, first tier, second tier, precision-recall curve, F1 score, discounted cumulative gain (DCG), and average normalized modified retrieval rank (ANMRR)~\cite{zhao2014affective,yao2020apse}.

We employ accuracy for emotion classification, mean squared error (MSE) for emotion regression, ANMRR for retrieval, and KL divergence for distribution learning. For accuracy, the larger the better; while for MSE, ANMRR, and KL divergence, smaller values indicate better results.


\subsection{Supervised Learning Results}
\label{sec:Supervised}

For emotion classification and regression, we compare the following methods:
\begin{itemize}
\item Traditional methods: principles-of-art based emotion features (PAEF)~\cite{zhao2014exploring}, adjective noun pairs (ANP) with SentiBank~\cite{borth2013large}, pretrained AlexNet~\cite{krizhevsky2012imagenet}, VGG-16~\cite{simonyan2015very}, and ResNet-101~\cite{he2016deep}. Support vector machine (SVM) or regression (SVR) with a radial basis function (RBF) kernel is used as the learning model.
\item Deep methods: fine-tuned (FT) AlexNet, VGG-16, and ResNet-101, MldrNet~\cite{rao2020learning}, SentiNet-A~\cite{song2018boosting}, WSCNet~\cite{she2020wscnet}, and PDANet~\cite{zhao2019pdanet}.
\end{itemize}

\begin{figure}[tb]
\begin{center}
\subfigure[Classification]{
\includegraphics[width=0.48\textwidth]{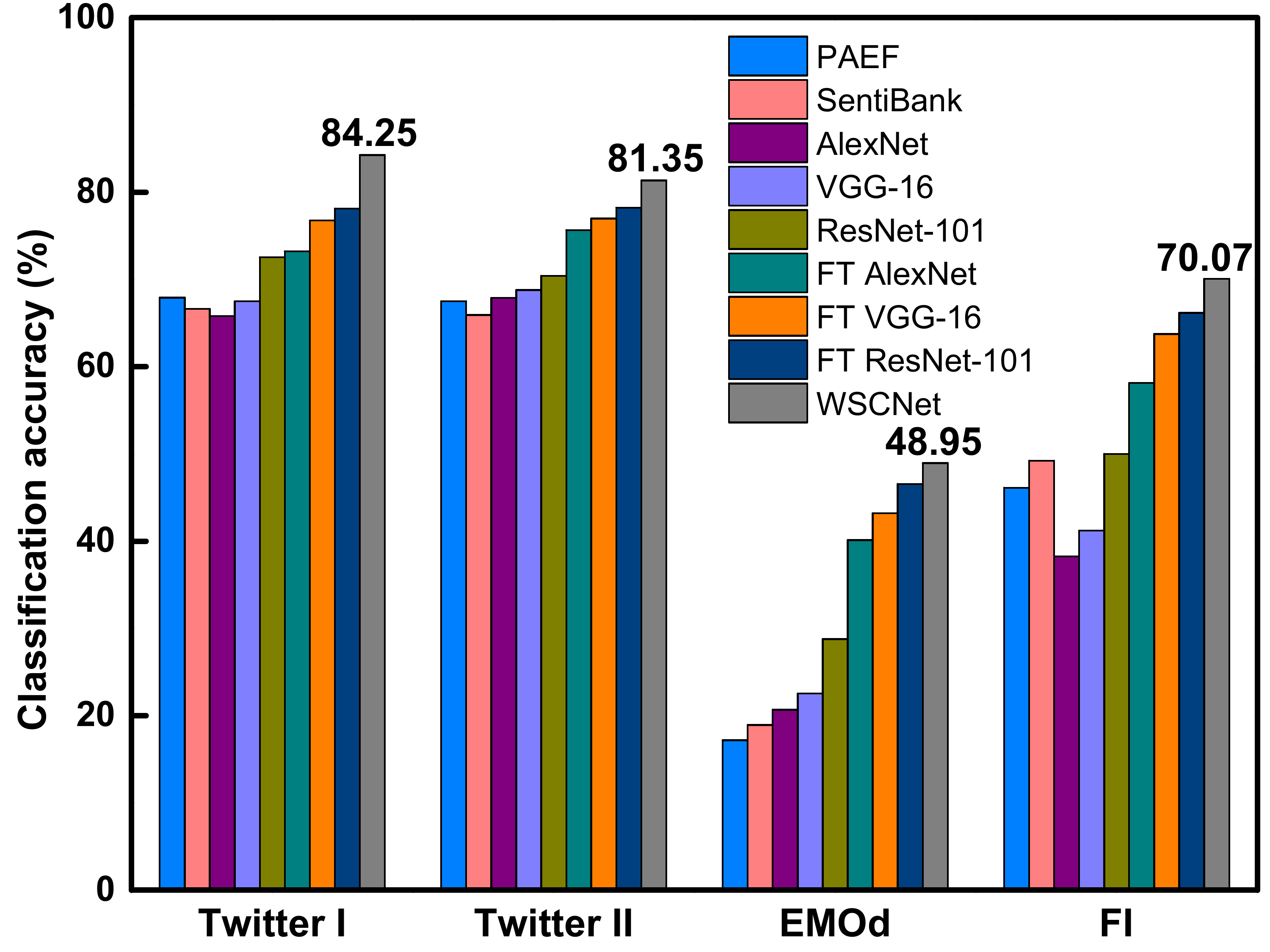}
\label{fig:Classification}
}
\subfigure[Regression]{
\includegraphics[width=0.48\textwidth]{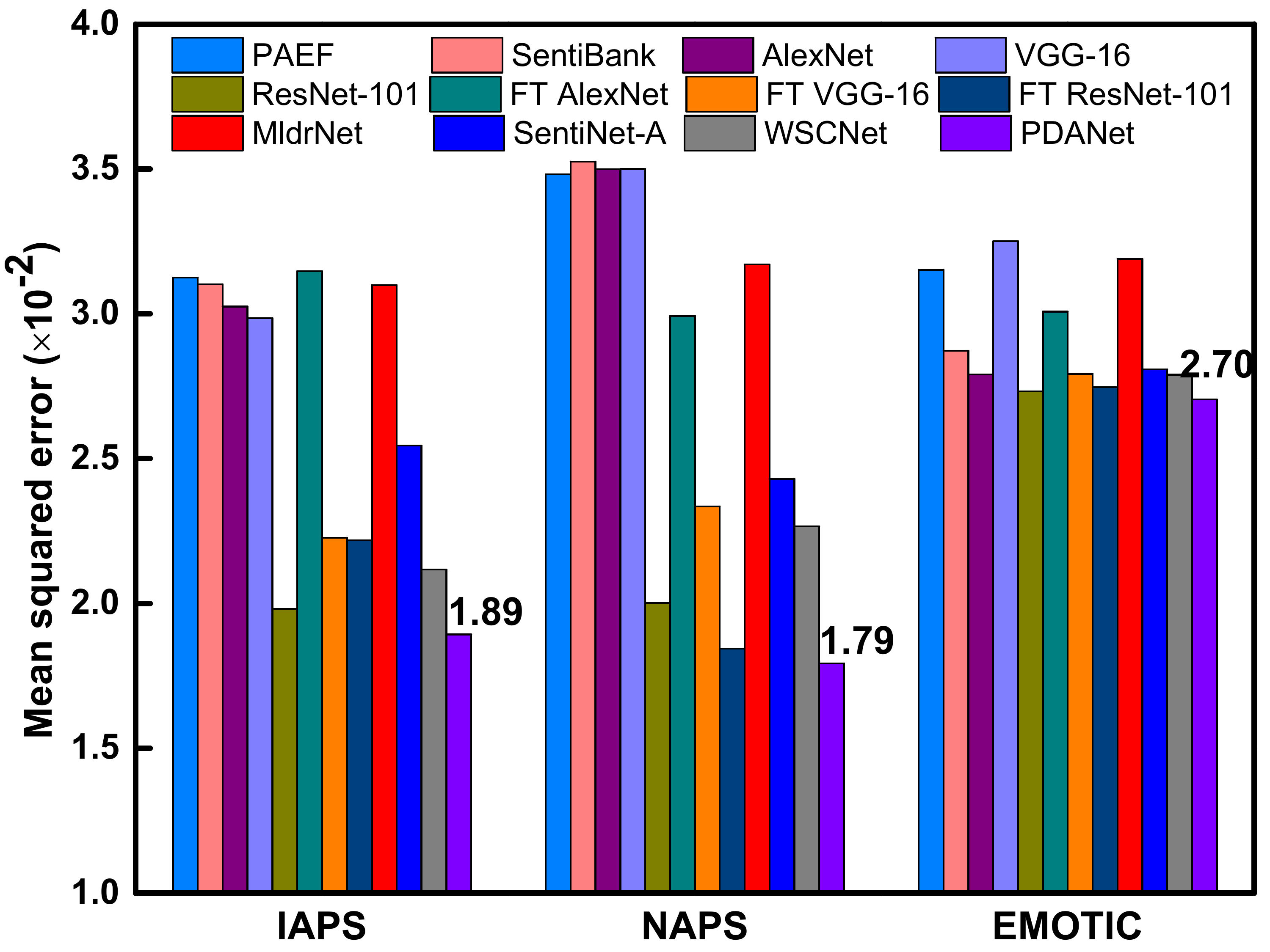}
\label{fig:Regression}
}
\subfigure[Retrieval]{
\includegraphics[width=0.48\textwidth]{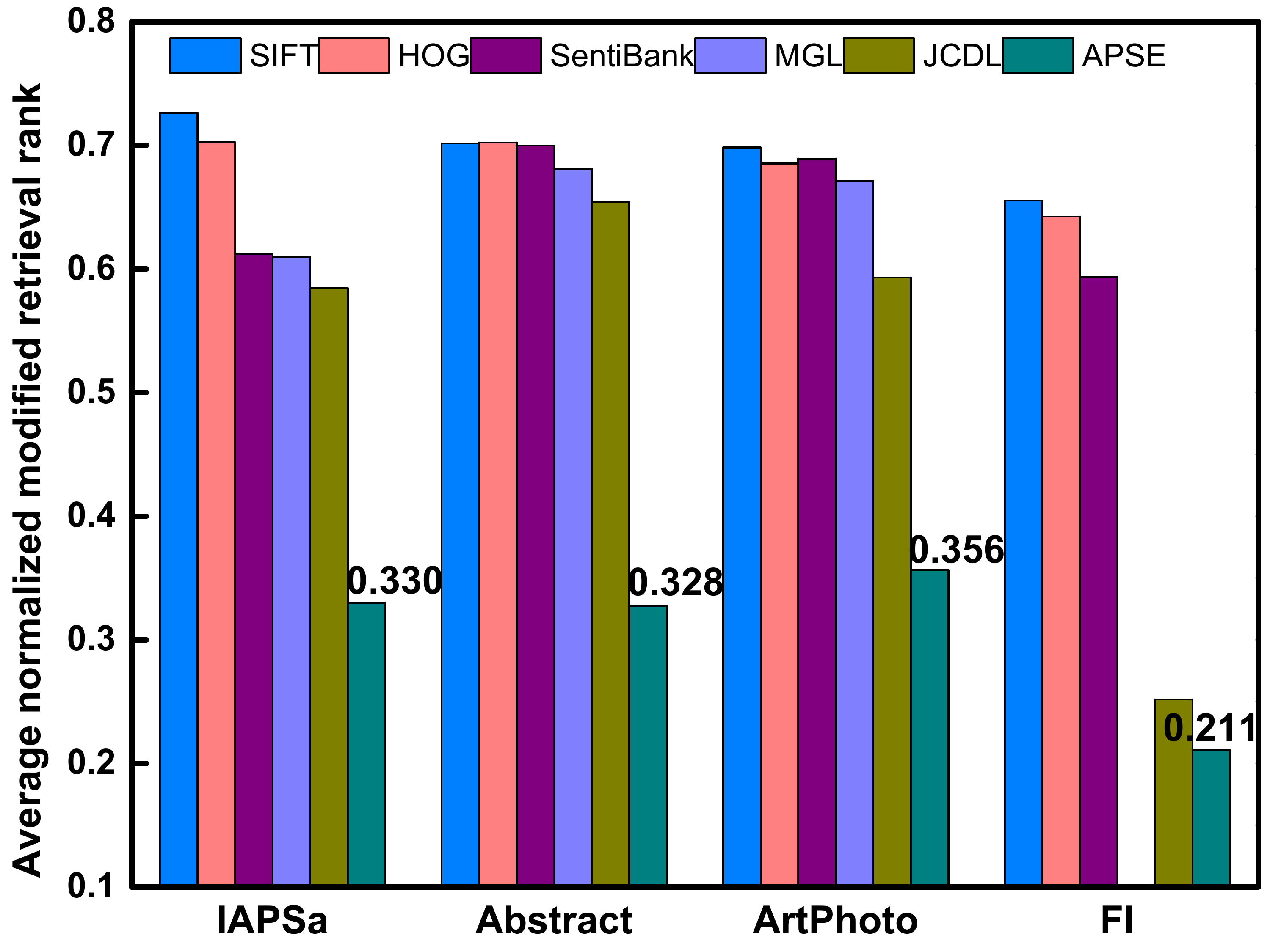}
\label{fig:Retrieval}
}
\subfigure[Distribution learning]{
\includegraphics[width=0.48\textwidth]{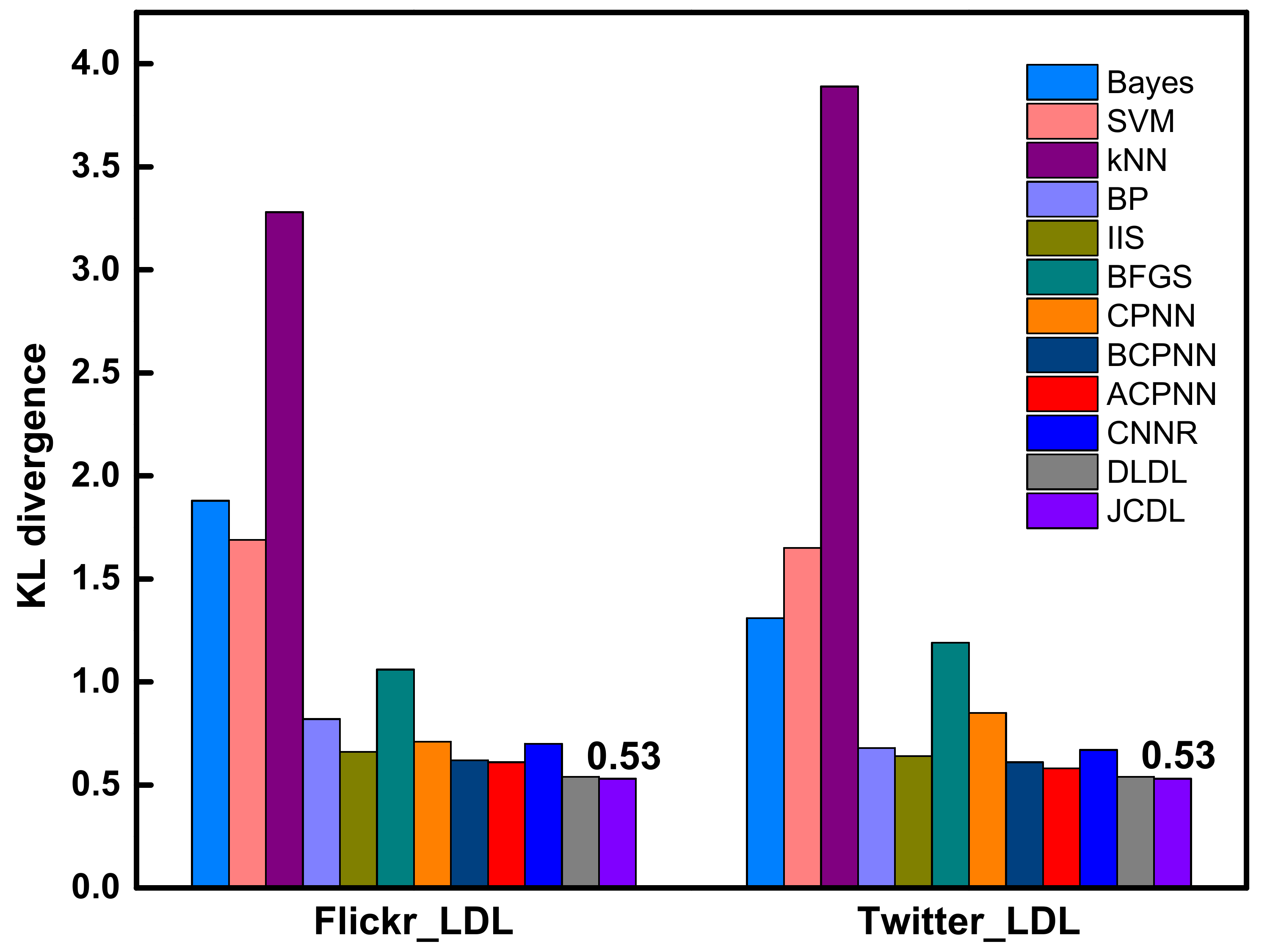}
\label{fig:Distribution}
}
\caption{Performance comparison of supervised learning methods for different IEA tasks, i.\,e., emotion classification, regression, retrieval, and distribution learning.}
\label{fig:Supervised}
\end{center}
\end{figure}

For emotion retrieval, we compare the performance of the following methods: SIFT~\cite{lowe1999object}, HOG~\cite{dalal2005histograms}, SentiBank~\cite{borth2013large}, Multi-graph learning (MGL)~\cite{zhao2014affective}, JCDL~\cite{yang2017joint}, and APSE~\cite{yao2020apse}. 

For emotion distribution learning, the compared methods include: Bayes, SVM, kNN, BP, IIS, BFGS, CPNN~\cite{geng2013facial}, BCPNN, 
ACPNN~\cite{yang2017learning}, CNNR~\cite{peng2015mixed}, DLDL~\cite{gao2017deep}, and JCDL~\cite{yang2017joint}.

\begin{figure*}[t]
\begin{center}
\centering \includegraphics[width=0.9\linewidth]{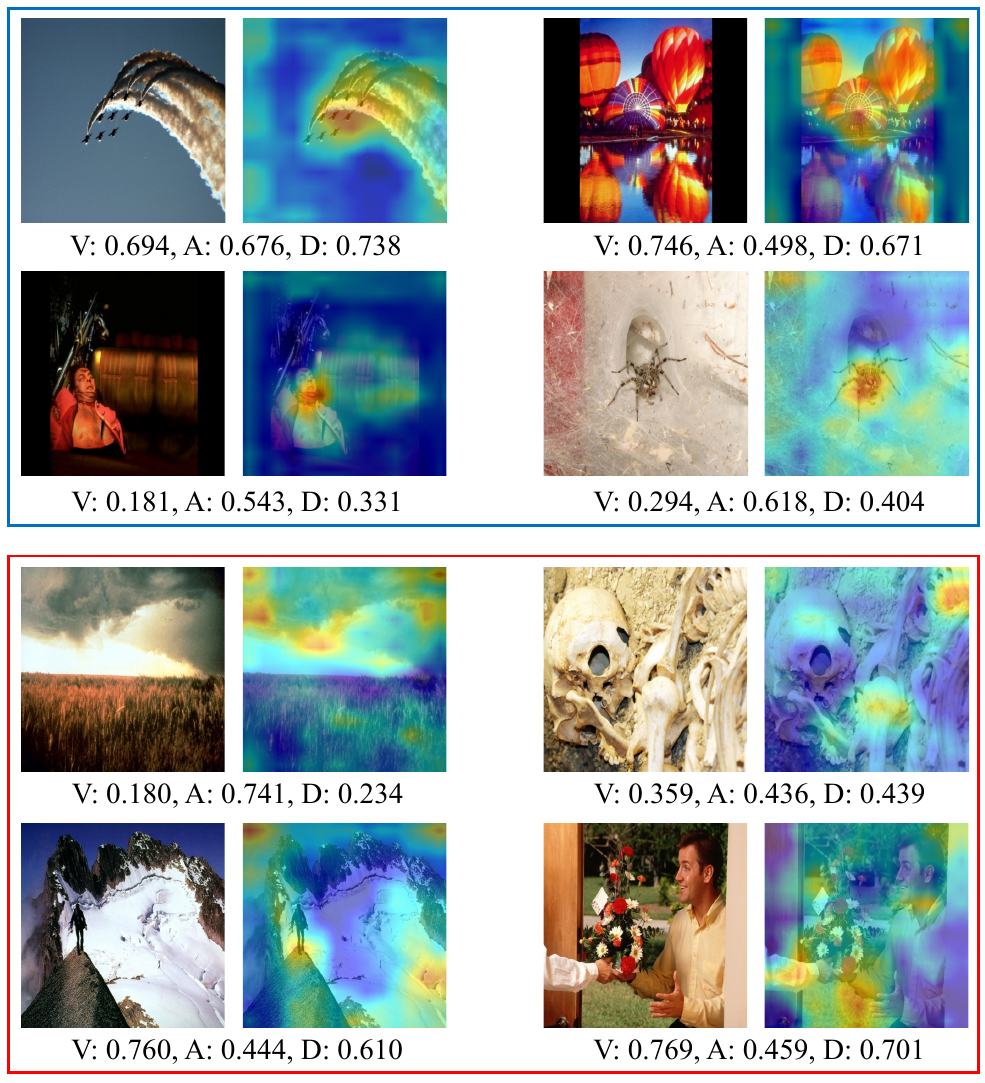}
\caption{Visualization of the learned attention maps by PDANet~\cite{zhao2019pdanet}. From left to right in each image pair are: original image from the test set and the combination of image and heat map. The ground truth VAD values are shown below each pair. Red regions indicate more attention. The attention in the above four examples in the blue rectangle can focus on the salient and discriminative regions, while the below in the red rectangle are failure cases.}
\label{fig:VisualizationPDANet}
\end{center}
\end{figure*}

The results of the above compared methods on emotion classification, regression, retrieval, and distribution learning are shown in Fig.~\ref{fig:Supervised}. From these results, we can conclude that:
\begin{enumerate}[~~~(1)]
\item Traditional hand-crafted low-level features in computer vision, such as SIFT and HOG, do not perform well on IEA tasks. For example, in Fig.~\ref{fig:Supervised} (c), the retrieval performance of SentiBank is much better than SIFT and HOG on the IAPSa dataset. 
\item Pretraind CNN features, especially the ones extracted from deep models (e.\,g., ResNet-101), achieve comparable and even better results as compared to hand-crafted specific features, such as PAEF and SentiBank, which demonstrates the generalization ability of deep features to new applications. For example, in Fig.~\ref{fig:Supervised} (a), the pretrained ResNet-101 features achieve 4.63\% and 5.92\%
performance gains on the Twitter I dataset for emotion classification as compared to PAEF and SentiBank.
\item Generally, fine-tuned deep models perform better than pretrained models. This is reasonable, since the pretrained models do not consider the specific characteristics of emotion-related features, while fine-tuned deep models can learn to adapt to the emotion datasets.
\item Deeper models usually perform better, which can be clearly observed when comparing AlexNet and ResNet-101 in Fig.~\ref{fig:Supervised} (a) and (b).
\item Specially designed models perform the best, such as APSE in Fig.~\ref{fig:Retrieval} and PDANet in Fig.~\ref{fig:Regression}; by modeling the specific characteristics of emotion, such as polarity-emotion hierarchy and attention mechanisms, these method can better bridge the affective gap.
\end{enumerate}

We visualize the learned attention of PDANet~\cite{zhao2019pdanet} using the heat map generated by the Grad-Cam algorithm~\cite{gradcam2017iccv} to show the model's interpretability. The results are shown in Fig.~\ref{fig:VisualizationPDANet}. More results on other visualizations can be found in our papers~\cite{yang2017joint,zhao2019pdanet,she2020wscnet,yao2020apse}. From the above four examples in the blue rectangle, we can see that PDANet can successfully focus on the salient and discriminative regions that determine the emotion of the whole image. For example, in the top right corner, the attention learned by PDANet focuses on the colorful balloons, which is strongly related to the positive emotion. We also show some failure cases in the red rectangle. As can be seen, for these cases, the background and foreground are difficult to be distinguished or the background is complex.

\begin{figure}[tb]
\begin{center}
\subfigure[Domain adaptation for classification]{
\includegraphics[width=0.48\textwidth]{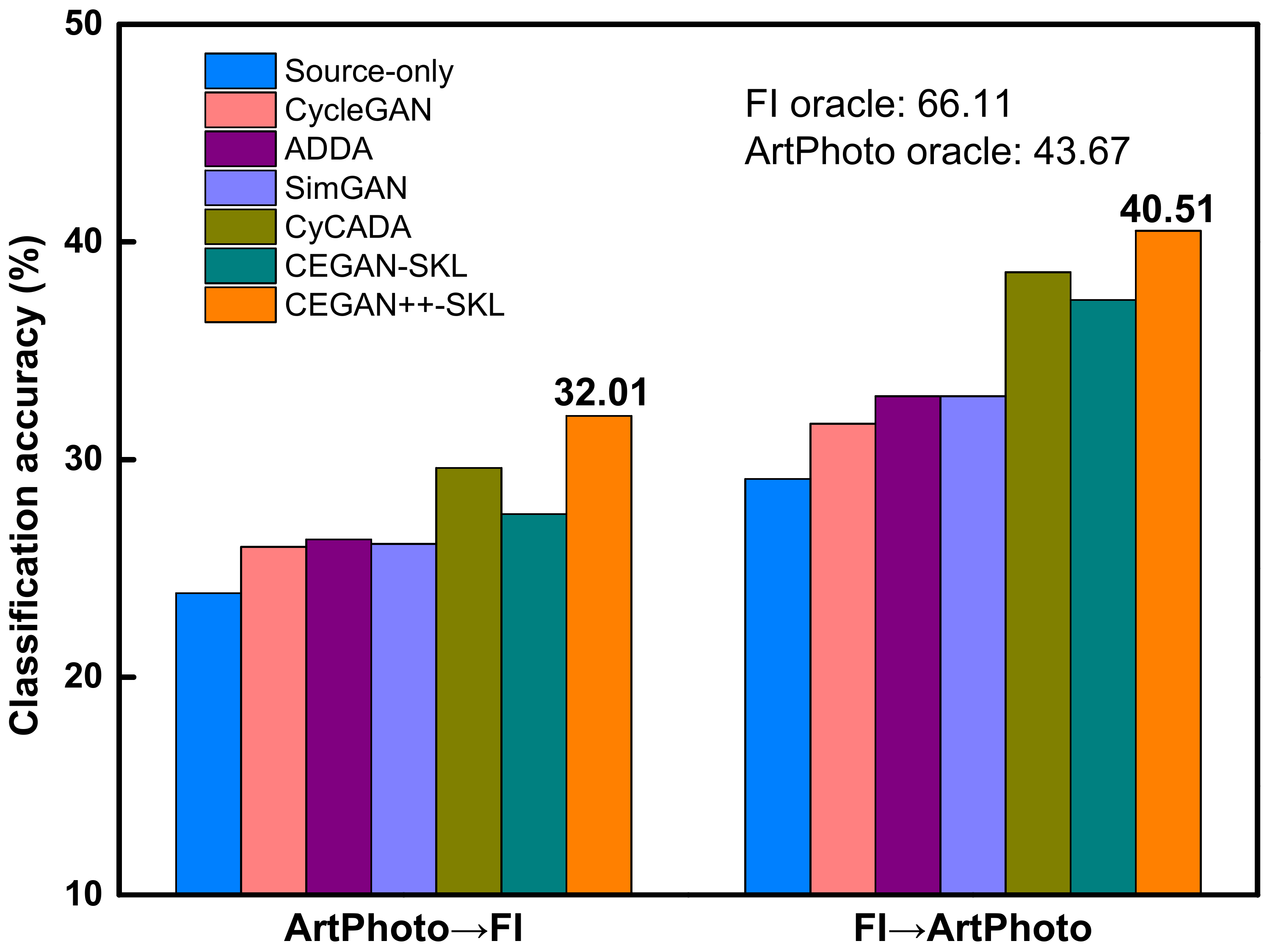}
\label{fig:ClassificationDA}
}
\subfigure[Domain adaptation for distribution learning]{
\includegraphics[width=0.48\textwidth]{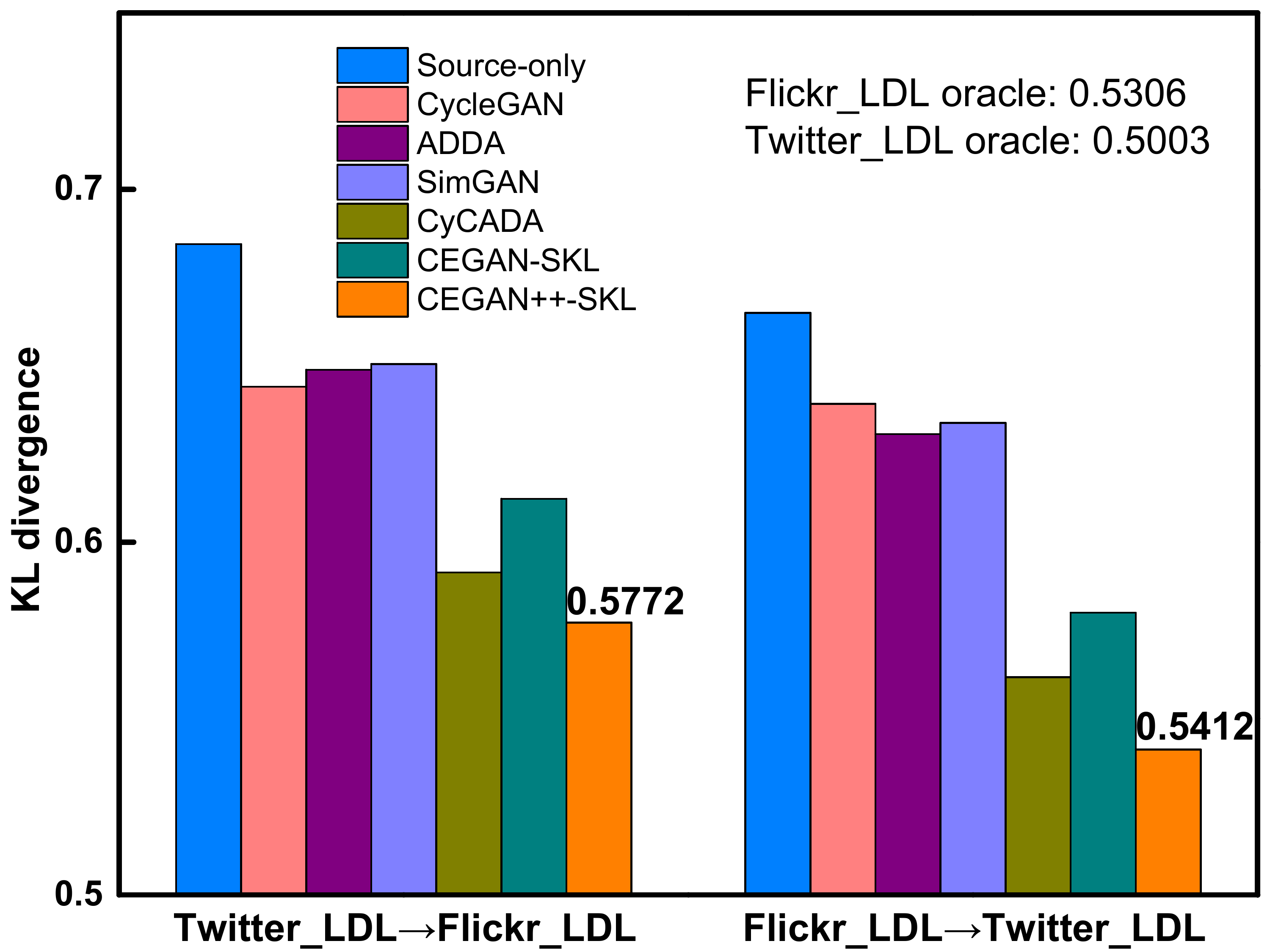}
\label{fig:DistributionDA}
}
\caption{Domain adaptation results for both emotion classification and distribution learning. For fair comparison and better visualization, the oracle results are shown in detailed numbers in the top right corner.}
\label{fig:IEADA}
\end{center}
\end{figure}

\subsection{Domain Adaptation Results}
\label{sec:DomainAdaptationResults}
For unsupervised domain adaptation for IEA, we report the performance comparison between CycleEmotionGAN++ (CEGAN++) with the following baselines: 
\begin{itemize}
\item Source-only: directly transferring the model trained on the source domain to the target domain;
\item Color style transfer methods: \textit{CycleGAN}~\cite{zhu2017unpaired};
\item UDA methods: \textit{ADDA}~\cite{tzeng2017adversarial}, \textit{SimGAN}~\cite{shrivastava2017learning}, and \textit{CyCADA}~\cite{hoffman2018CyCADA};
\item Oracle: training and testing on the target domain, which can be viewed as an upper bound.
\end{itemize}

\begin{figure*}[t]
\begin{center}
\centering \includegraphics[width=0.9\linewidth]{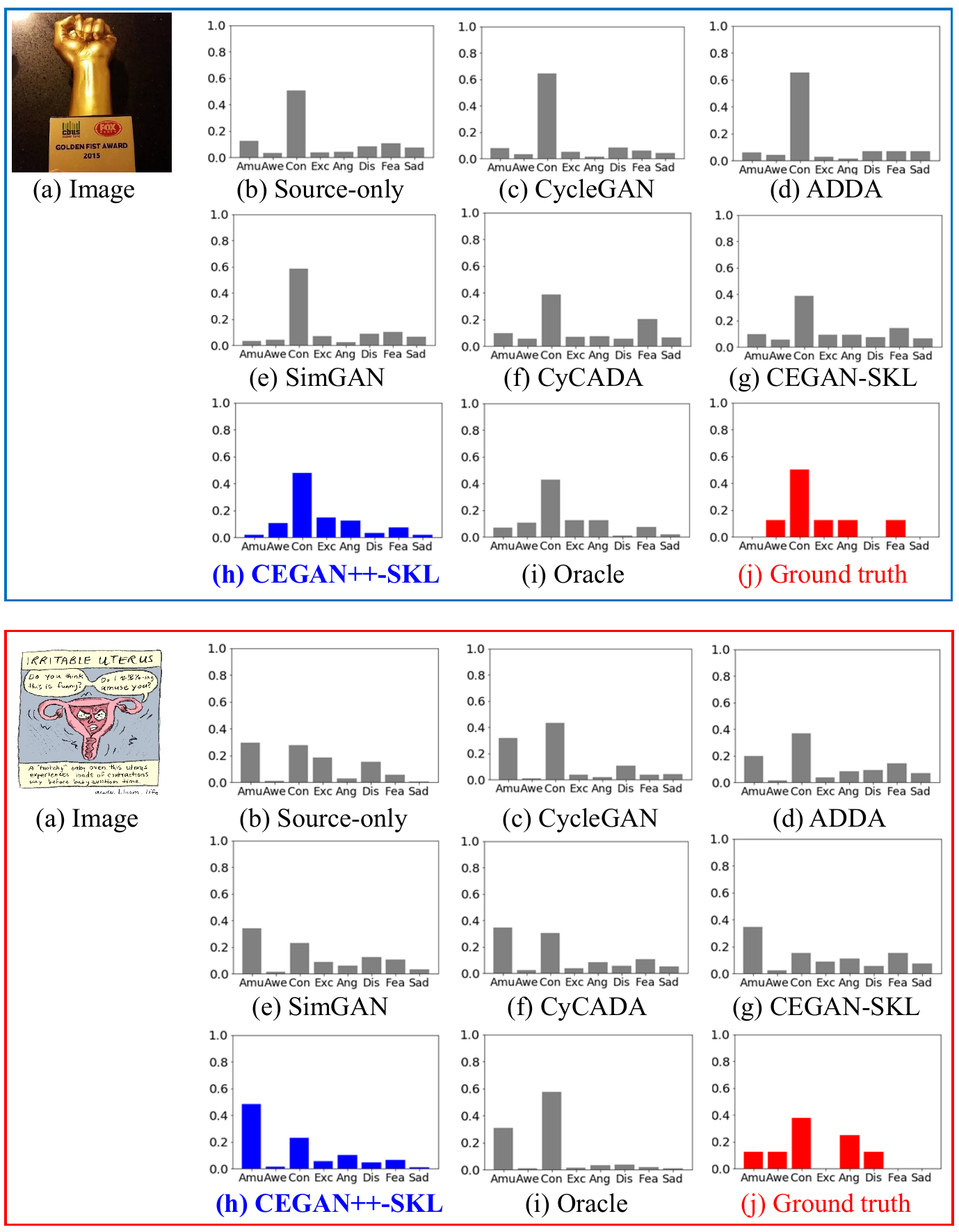}
\caption{Visualization of predicted emotion distributions on the Twitter-LDL dataset by  CycleEmotionGAN++-SKL (CEGAN++-SKL)~\cite{zhao2021emotional} and several other baselines. In the above example, CEGAN++-SKL can predict similar emotion distribution to the ground truth; while the below example shows a failure case.}
\label{fig:twitterLDL}
\end{center}
\end{figure*}

The task classifiers use the ResNet-101~\cite{he2016deep} architecture pretrained on ImageNet. Please see~\cite{zhao2021emotional} for more implementation details. The performance comparisons between CEGAN++ and the above-mentioned approaches are shown in Fig.~\ref{fig:IEADA}. From the results, we can observe that:
\begin{enumerate}[~~~(1)]
\item Because of the influence of domain shift, directly transferring the models trained on the source domain to the target domain does not perform well. For example, when adapting from ArtPhoto to FI, i.\,e., training on ArtPhoto and directly testing on FI, the classification accuracy is only 23.86\,\%. The model's low transferability from one domain to another motivates the necessity of domain adaptation research.
\item CEGAN++ achieves the best result among all domain adaptation methods for both emotion classification and distribution learning. The superiority of CEGAN++ for adapting image emotions benefits from the following aspects: pixel-level and feature-level alignments to align the source and target domains, dynamic emotional semantic consistency to dynamically preserve the emotion information before and after image translation.
\item There is still an obvious gap between all the domain adaptation methods and the oracle setting that is trained on the target domain. For example, the oracle accuracy on FI is 66.11\,\%, and the best adaptation result is 32.01\,\%. Future efforts are still needed to further bridge the domain shift between different domains. 
\end{enumerate}

Fig.~\ref{fig:twitterLDL} shows some predicted emotion distributions by different domain adaptation methods on the Twitter-LDL dataset, including one successful example and one failure case. More visualization results can be found in~\cite{zhao2021emotional}. From the above example, we can clearly see that the predicted emotion distribution by CEGAN++ is close to the ground truth distribution, which demonstrates its effectiveness for visual emotion adaptation. In the below failure case, we can see that even the oracle does not perform well, which indicates the challenges of IEA, requiring further research efforts.

\section{Conclusions and Future Research Directions}
\label{Sec:Conclusion}
We introduced recent advances on image emotion analysis (IEA) from different aspects with the focus on our recent efforts. First, we summarized related psychological studies to understand how emotion is measured. Second, based on the emotion representation models, we defined the key computational problems and widely used supervised frameworks, and then we introduced three major challenges in IEA. Third, we summarized and compared representative methods on emotion feature extraction and learning methods for different IEA tasks. Finally, we briefly described existing datasets and presented an experiment with some of the current state-of-the-art approaches. 

Although much research attention has been paid to IEA with promising methods proposed, the overall performance is still not perfect and there is still no solution commonly accepted to address these problems. Many issues in IEA are still open and deserve our further research efforts. We do believe with the progress of multiple disciplines, such as psychology, brain science, and machine learning, IEA will continue to be a hot research topic. At the end, we provide some topics that are well worth considering and investigating.

\textbf{Context-aware Image Emotion Analysis.}
Besides extracting discriminative visual features, incorporating available context information can also contribute to the IEA task~\cite{kosti2020context}. (1) \textit{Image context}. Similar image content in different contexts might induce totally different emotions, either within an image or across modalities. For example, if we see some soldiers smiling surrounded by flowers, we may feel moved for their contributions to the nation, such as epidemic fighting; but if there is a nearby dead child, we may feel angry for their atrocity. If we see a famous football player crying on his knees, the audience might feel sad; but if this is after winning a game, the audience especially the team's amateurs my feel excited. (2) \textit{Viewer context}. The context in which a viewer is watching an image and the viewers' prior knowledge (e.\,g., personality, gender, and culture background) can also contribute a lot to the emotion perception. For example, a viewer's current emotion might be strongly correlated with his/her recent past emotions~\cite{zhao2018predicting}. (3) \textit{Image-viewer interaction}. Humans' emotion perception is a complex process involving both the stimulus and the physical and psychological changes. Combining such implicit and explicit channels are helpful in the final IEA performances.

\textbf{Determining Intrinsic Emotion Features and Localizing Image Emotions to Image Regions.}
As shown in~\cite{zhao2014affective}, the emotions of different kinds of images are determined by different features. If we can firstly know the image type, we can select corresponding features that are discriminative for IEA. But what image types should we define for emotion prediction is still unclear. Attempting large scale data-driven approaches is worth trying. Although deep learning based methods achieve promising results for IEA, the explainability on why these methods work, i.\,e., what features they focus on, has not been fully investigated. Determining the intrinsic features to understand what makes an image amusing, sad or frightening still remains an open problem.

Sometimes, the emotion of an image is determined by the overall appearance of the image. Occasionally, the emotion is reflected by some key image regions. It would be helpful for us to localize these key regions, which can be changed or replaced to change the image emotions~\cite{peng2014framework}. We can use traditional segmentation methods to segment images into regions and recognize the emotions of each region. Or we can train classifiers to detect the key regions. For example, ANP classifiers are trained hierarchically to localize objects~\cite{chen2014object}. More recent emotional region localization methods are based on attention~\cite{zhao2019pdanet} and sentiment maps~\cite{she2020wscnet}. Besides an emotion classification branch, WSCNet trains another weakly-supervised detection branch to learn the sentiment specific soft map by a fully convolutional network with the cross spatial pooling strategy~\cite{she2020wscnet}. PDANet jointly considers the spatial and channel-wise attention through which we can obtain the attentive and discriminative regions~\cite{zhao2019pdanet}. Jointly combining the advantages of traditional object detection methods and the characteristics of image emotions might motivate new solutions.

\textbf{Understanding Emotions of 3D Data.}
Most existing works on emotion and sentiment analysis of general images are based on 2D images. But with the wide popularity and public use of somatosensory equipment such as Kinect, more and more 3D data (e.\,g., 2D images and depth) are created and shared just like personal photos and web videos. Compared with traditional intensity and color images, 3D data contain more information and have several advantages, such as being useful in low light levels and being color and texture invariant~\cite{shotton2011real}. Some research efforts have been dedicated to recognizing 3D facial expressions~\cite{sandbach2012recognition}. However, few works on generalized 3D emotion analysis have been published. To the best of our knowledge, no public emotion dataset of general 3D data is released. Building a large scale 3D emotion dataset is an urgent need and of great value. Using social network data may help to reduce the time-consuming and tedious labelling task. With the rapid development of 3D content analysis, understanding the emotions of 3D data will become a hot research topic.

\textbf{Image Emotion Analysis in The Wild.}
Existing IEA methods are mainly based on specific settings, such as training on small datasets with limited annotators. However, in real-world applications, the IEA problems are much more complex and difficult. For example, the given datasets might contain inaccurate annotations and much noise that is unrelated to emotion; training data is given incrementally and the emotion categories are becoming more fine-grained gradually; the labeled data is unbalanced across different emotion categories; the test set has different styles from the training set; only limited computing resource is available. How to design an effective and efficient IEA model that can still work under these practical settings is still open.

\textbf{Novel and Real-world Applications Based on IEA.}
Due to the relatively limited progress in the early years, e.\,g., low performance, emotion has not been widely deployed in real applications. With recent development of deep learning and large-scale datasets, the IEA performance has been and will continue to be significantly boosted. Therefore, we foresee an emotional intelligence era in the near future with many novel and real-world IEA-based applications. For example, we can understand how artists express emotions through their artworks and use the learned principles in painting education. In fashion advertisement, we can design the best matching between clothes and models to attract users' attention and improve user experience, which can lead to increasing sales.

\textbf{Security, Privacy, and Ethics of IEA.} 
As discussed above, viewers' prior knowledge, such as identity, age, and gender, can contribute to the IEA performance. However, this information is confidential, which should not be shared or leaked. Therefore, protecting the security and privacy must be taken into account in real applications. Further, there is no related law regarding the IEA tasks, especially for personalized scenarios. People might not want their emotion to be recognized and used. From the perspective of ethics, it is important to consider such impact, which requires the joint efforts from different communities, such as psychology, cognitive sciences, and computer science.


\bibliographystyle{spmpsci}
\bibliography{visualEmotion}

\end{document}